\documentclass{article}

\usepackage{PRIMEarxiv}

\usepackage[utf8]{inputenc} % allow utf-8 input
\usepackage[T1]{fontenc}    % use 8-bit T1 fonts
\usepackage{hyperref}       % hyperlinks
\usepackage{url}            % simple URL typesetting
\usepackage{booktabs}       % professional-quality tables
\usepackage{amsfonts}       % blackboard math symbols
\usepackage{nicefrac}       % compact symbols for 1/2, etc.
\usepackage{microtype}      % microtypography
\usepackage{fancyhdr}       % header
\usepackage{graphicx} 
\usepackage{subcaption}% graphics
\usepackage{arydshln}
\graphicspath{{media/}}     % organize your images and other figures under media/ folder

\title{Towards AI-powered Language Assessment Tools for Dementia
}

\author{
  Mahboobeh Parsapoor (Mah Parsa) \\
 CRIM \\
  Montreal\\
  \texttt{\ mah.parsa@crim.ca} \\
  \AND
  Muhammad Raisul Alam \\
  Department Occupational Science and Occupational Therapy\\
  University of Toronto \\
  Toronto\\
  \texttt{raisul.alam@utoronto.ca} \\
  \AND
  Alex Mihailidis\\
  Department Occupational Science and Occupational Therapy\\ University of Toronto\\
  Toronto\\
 \texttt{ alex.mihailidis@utoronto.ca} \\
}

\begin{document}
\maketitle
\begin{abstract}
The main objective of this paper is to propose an approach for developing an \textit{Artificial Intelligence}~(AI)-powered \textit{Language Assessment}~(LA) tool. Such tools can be used to assess language impairments associated with dementia in older adults. The \textit{Machine Learning}~(ML) classifiers are the main parts of our proposed approach, therefore to develop an accurate tool with high sensitivity and specificity, we consider different binary classifiers and evaluate their performances. We also assess the reliability and validity of our approach by comparing the impact of different types of language tasks, features, and recording media on the performance of ML classifiers. 
Our approach includes the following steps: 1) Collecting language datasets or getting access to available language datasets; 2) Extracting linguistic and acoustic features from subjects' speeches which have been collected from subjects with dementia ($N$=9) and subjects without dementia ($N$=13); 3) Selecting most informative features and using them to train ML classifiers; and 4) Evaluating the performance of classifiers on distinguishing subjects with dementia from subjects without dementia and select the most accurate classifier to be the basis of the AI tool.
%\newline
Our results indicate that 1) we can find more predictive linguistic markers to distinguish language impairment associated with dementia from 
participants' speech produced during the~\textit{Picture Description} (PD) language task than the~\textit{Story Recall} (SR) task; and 2) phone-based recording interfaces provide more high-quality language datasets than the web-based recording systems
Our results verify that the tree-based classifiers, which have been trained using the linguistic and acoustic features extracted from interviews' transcript and audio, can be used to develop an AI-powered language assessment tool for detecting language impairment associated with dementia.
\end{abstract}

% keywords can be removed

\section{Introduction}
More than 50 million people worldwide are living with different types of neurodegenerative dementias including \textit{Alzheimer's Disease}~(AD), Vascular Dementia, Lewy Body Dementia, and Frontotemporal Lobar Dementia ~\cite{ripich2004neurodegenerative}. These are among the leading global neurodegenerative diseases and have notable economic impacts on individuals and societies~\cite{nichols2019global}. To mitigate the impact of neurodegenerative dementias on older adults and help them plan for the future~\cite{santacruz2001early}, early detection of dementia is necessary. It would help older adults at the early stages of the disease seek out different intervention programs~\cite{green1997early}, including psycho-social interventions (e.g., walking programs and art therapy)~\cite{duan2018psychosocial}, non-pharmaceutical intervention programs (e.g., music interventions~\cite{fischer2019music}) as well as clinical interventions so that they can maintain their quality of life~\cite{logsdon2007evidence} at the normal level and slow down disease progression.

There is no single test to diagnose dementia; clinicians run different tests, including cognitive \footnote{\footnotesize{The cognitive tests evaluate cognitive function including memory, orientation, attention, reasoning and judgment, language skills, and attention}}~\cite{10.1001/jamainternmed.2015.2152,creavin2016mini,10.3389/fpsyt.2019.00878},  neuropsychological, neurological, brain-imaging, laboratory tests, and psychiatric evaluation to detect cognitive impairment associated with dementia~\cite{daly2005initial}. 
Two well-known cognitive assessment tools are~\textit{Montreal Cognitive Assessment}~(MoCA) and the~\textit{Mini-Mental State Examination}~(MMSE)~\cite{kalish2016mini, nasreddine2005montreal}. The MoCA test consists of a 30-points scales~\cite{kalish2016mini} to identify subjects with AD and~\textit{Mild Cognitive Impairment}~(MCI)~\footnote{\footnotesize{MCI refers to the condition where an older adult experiences cognitive impairment especially in tasks related to orientation and judgment~\cite{chiu2019nmd}}}~\cite{nasreddine2005montreal}; while, the MMSE includes 11 questions \cite{10.1001/jamainternmed.2015.2152} to assess impairments related to five cognitive functions such as orientation, attention, memory, language and visual-spatial skills \cite{10.1001/jamainternmed.2015.2152}. The MoCA and MMSE assessment tools 
are quick and cost-effective, and widely used by clinicians and psychiatrists \cite{o2005use, vertesi2001standardized}. 
However, these tests cannot diagnose dementia with high sensitivity and specificity~\cite{sheehan2012assessment, kalish2016mini, nasreddine2005montreal}. This can be problematic since the false results of such tests can significantly impact health insurance and some social rights of the individuals~\cite{chambers2017dementia}. 
Therefore, to detect dementia quickly and effectively, it has been suggested that such tests be considered as the beginning steps in early detection programs and can be combined with the~\textit{Language Assessment}~(LA) tools~
\cite{chaves2011cognitive, sheehan2012assessment}. 

Clinical services and psychiatrists can detect dementia patients using LA tools, which have been recognized as low-cost and effective tools with high specificity and sensitivity of diagnosis~\cite{el2018novel} at the earliest stage of the disease. These tools can detect language impairments, which are signs of the first cognitive manifestations of any types of dementia, specifically the onset of AD~\cite{klimova2015alzheimer} and MCI. 
LA tools can also be useful to identify different types of language impairment including difficulties with finding a relative expression, naming, and word comprehension and various level of language impairments ~\cite{klimova2015alzheimer}. For example, the tools can identify 1) lexical-semantic language problems such as naming the things or being vague in what they want to say~\cite{klimova2015alzheimer}; 2) signs of empty speech (e.g., ``The thing is over there, you know");  3) phonological, morphological, syntactical problems and the lack of verbal fluency; which are patients' language problems at the mild, moderate and severe stage of AD respectively. Thus, the tools make possible to clinicians to detect different types of dementia and its various stages.

Furthermore, these tools can detect language disorders, incoherent speech, tangentiality and grammatical error, lexical retrieval difficulties, auditory comprehension difficulties, grammatical and spelling failures~\cite{doi:10.1111/hsc.12887,green2018investigating, szatloczki2015speaking} in the subjects. These signs are associated to \textit{Language and Communication Impairment} in patients with dementia. In particular, LA tools that analyze spontaneous speech, produced during the completion of cognitive tasks~\cite{ferreira2011neuroimaging} can recognize linguistic features associated with language performance deficits in elderly individuals \cite{mccullough2019language}. Therefore, they are efficient methods to diagnose AD/MCI in elderly adults \cite{vestal2006efficacy}. 

As mentioned earlier, the main advantage of using LA tools is their cost-effectiveness and user-friendliness. It is beneficial to patients and clinicians alike to use them. 
Thus, we propose an approach for developing an AI-powered language assessment tool to detect dementia. Unlike the previous research papers~\cite{godino2005support, guinn2012language, orimaye-etal-2014-learning, asgari2017predicting, karlekar2018detecting}, we have not just focused on examining classifiers to distinguish subjects with dementia from subjects without dementia, rather we have defined different experiments to understand the impact of the language tasks, types of features and recording media on the efficiency of the tool.  More specifically, we seek to 
find out the impact of 1) different language tasks, e.g., the picture description and the story recall tasks, 2) recording media, e.g., phone vs web-based interfaces, and 3) linguistic and acoustic features on the efficiency AI-powered language assessment tools. 

Another contribution of the paper is that we  have introduced four metrics to measure incoherence and tangential speech in elderly individuals. 

The remainder of the paper is organized as follows. Section 2 provides a brief overview of using \textit{Machine Learning} (ML) to develop language assessment tools. 
Section 3 describes our approach to develop AI-powered LA tools. Section 4 presents our results.
Section 5 discusses the data limitation, feature selection, validity, reliability, fairness, and explainability aspects of LA tools. Finally, Section 6 concludes the article highlighting its main contributions and our future direction.

\section {Related Works}
Detecting language impairments using ML has captured the attention of researchers in the field of neurodegenerative disease \cite{godino2005support, guinn2012language, orimaye-etal-2014-learning, asgari2017predicting, karlekar2018detecting}. We generally combine the following steps to develop AI-based LA tools:
1) Collecting language datasets or getting access to available language datasets;  
2) Feature Engineering: i) Extracting linguistic and acoustic features; 
ii) Employing various feature selection methods to select informative features;
3) Training different classifiers using multiple sets of features, and selecting ML algorithms with the highest performance. % to be the assessment tool's basis.
In this section, we describe the research related to each of the steps.
 
\subsection{Language Datasets}
\label{Language Datasets}
\begin{figure}[!h]
    \centering
    {\includegraphics[scale=.4]{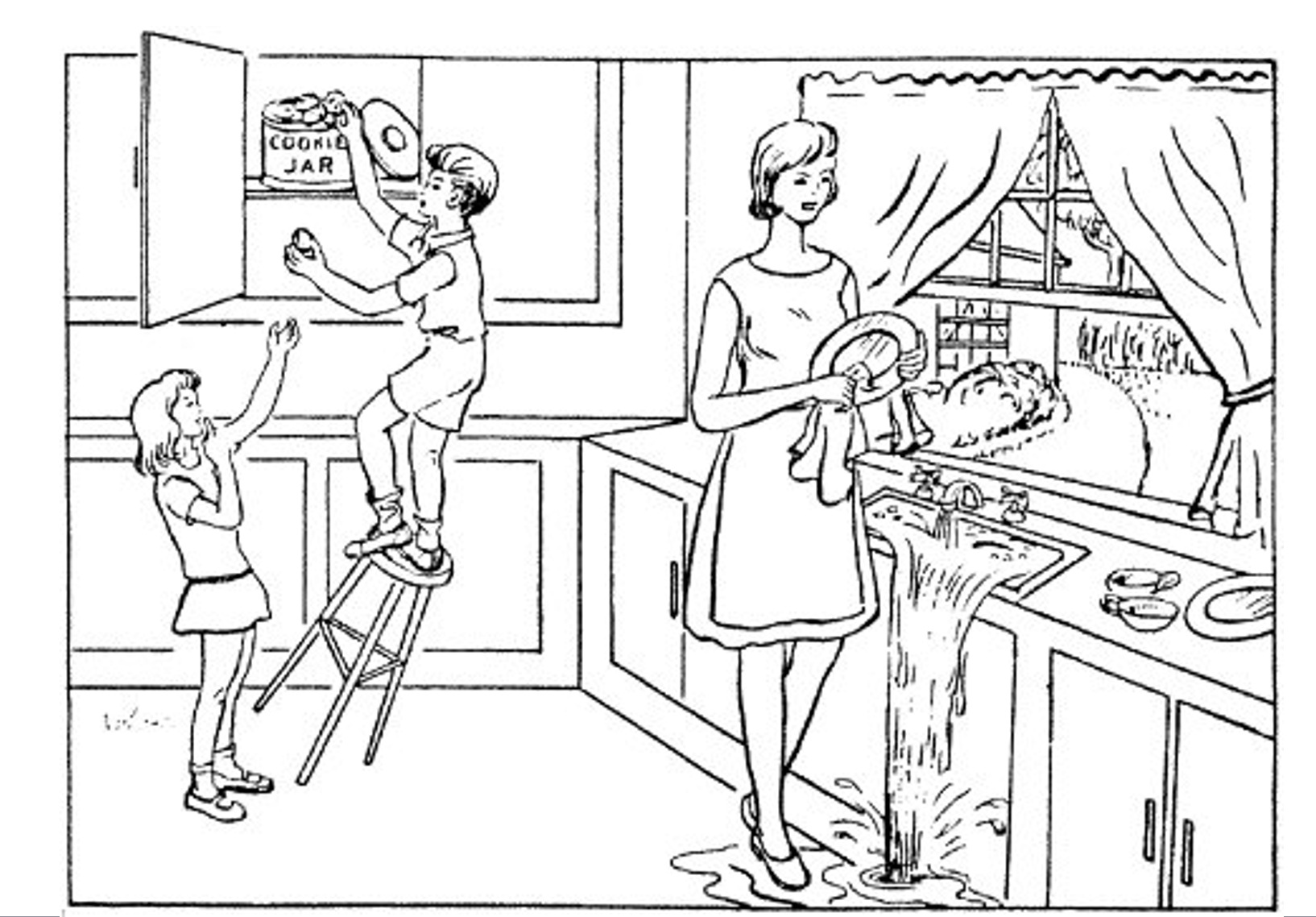}}
    \caption{The Cookie Theft Picture from the Boston Diagnostic Aphasia Examination. 
    For the PD task, the examiner asks subjects to describe the picture by saying: \textbf{Tell me everything you see going on in this picture". 
Then subjects might say, "there is a mother who is drying dishes next to the sink in the kitchen. She is not paying attention and has left the tap on. As a result, water is overflowing from the sink. Meanwhile, two children are attempting to make cookies from a jar when their mother is not looking. One of the children, a boy, has climbed onto a stool to get up to the cupboard where the cookie jar is stored. The stool is rocking precariously. The other child, a girl, is standing next to the stool and has her hand 
outstretched ready to be given cookies.}\cite{cummings2019describing}}
    \label{fig:PD}
\end{figure}

\begin{figure}[!h]
    \centering
    {\includegraphics[scale=.6]{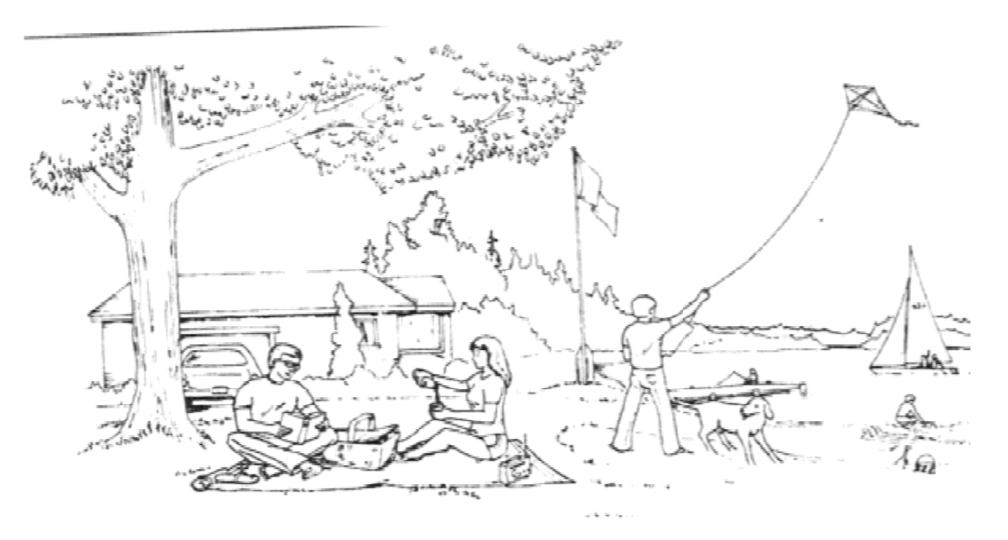}}
    \caption{The the picnic scene from The Arizona Alzheimer’s Disease Center (ADC) \cite{weissenbacher2016automatic}. A normal subject might say: A family outing at a lake shore showed people doing several things. Mom and Dad sat on a blanket while dad read a book. Dad was over comfortable without his shoes, while mom listened to the radio and poured herself a cup of coffee. Junior was having fun flying his kite, and the family dog was interested in what all was going on. An- other of the family was spending quiet time and fisher- man, and another was playing in the shallow water. Other friends waved to them as they sailed by. It was a perfect day with just enough wind to move the flag and provide lift for the kite. It must have been comfortable sitting under the shade tree.
    }
    \label{fig:PD_PS}
\end{figure}

We need labelled language datasets of older adults to develop supervised ML algorithms that can detect language impairments in patients. So far different language datasets, such as \textit{Carolina Conversations Collections}~(CCC)~\cite{pope2011finding} and  \textit{DementiaBank}~(DB) \footnote{\footnotesize{The DB dataset has been collected by recording the voice of patients with AD ($N$=167) and healthy control ($N$=97) while completing a picture description task.}}~\cite{becker1994natural} have been introduced. The datasets were obtained using various language tests such as the \textit{Boston Naming Test}~(BNT) which is a standard test to assess language performance in participants with aphasia or dementia. Deficits in naming production appear in the first stages of Alzheimer’s disease and boost with time. Thus, BNT is one of the tests that can be used to detect the disease and follow its course.
Moreover, it is useful in discriminating healthy elderly persons and those with dementia \cite{calero2002usefulness}. 
The language tests aim to collect data to assess various aspects of language impairment in subjects. 
For example, the \textit{Picture Description}~(PD) task is usually used to evaluate the semantic knowledge in subjects  ~\cite{slegers2018connected}. 
Using the Cookie Theft (see Figure \ref{fig:PD}) or the Picnic Scene (see Figure \ref{fig:PD_PS}) for the PD task, 
we can assess the structural language skills \cite{cummings2019describing} of patients and amplify signs of language impairment. 
On the other hand, the \textit{Story Recall}~(SR) task\footnote{\footnotesize{During the story recall task, participants are shown a short passage with one of the following options 1) My Grandfather, 2) Rainbow or 3) Limpy 
\href{https://itcdland.csumb.edu/~mimeyer/CST251/readingpassages.html}{that are three well-known passage to assess memory capacity of participants.} 
}} can help assess impairment in episodic and semantic memory and also global cognition.

\subsection{Feature Engineering}
\label{Feature Engineering}
One of the main steps to develop AI-based LA tools is to extract linguistic and acoustic features from raw text and audio files. Tables~\ref{table:tab1_L} and ~\ref{table:tab_acoustics_lit} present the lists of linguistic and acoustic features respectively. These extracted features (or a subset of these features) can directly be used to train ML  algorithms.%, or a set of informative features selected from them can be prepared to train ML algorithms. 

\begin{table}[!t]
\caption{\footnotesize{List of Linguistic Features reported in the Literature within AD Domain. Different linguistic features, including lexical, syntactic, semantic and pragmatic features can be associated with different types of language deficits in patients with AD and MCI~\cite{araujo2015linguagem}. }}
\label{table:tab1_L}
\setlength{\tabcolsep}{4pt}
\begin{tabular}{|p{150pt}|p{90pt}|p{120pt}|p{50pt}|}
\hline
Name&Type&Cognitive Function& References\\
\hline
Coordinated sentences&Syntactic& Syntactic processing&\cite{orimaye2017predicting}\\
Subordinated sentences&Syntactic& Syntactic processing&\cite{orimaye2017predicting}\\
Reduced sentences&Syntactic& Syntactic processing&\cite{orimaye2017predicting}\\
Number of predicates &Syntactic& Syntactic processing&\cite{orimaye2017predicting}\\
Average number of predicates&Syntactic& Syntactic processing&\cite{orimaye2017predicting}\\
Dependency distance&Syntactic& Syntactic processing&\cite{orimaye2017predicting}\\
Number of dependencies&Syntactic& Syntactic Processing&\cite{orimaye2017predicting}\\
Average dependencies per sentence &Syntactic& Syntactic processing&\cite{orimaye2017predicting}\\
Production rules&Syntactic& Syntactic processing&\cite{orimaye2017predicting}\\
Noun Rate (NR)&Syntactic& Cognitive strength&\cite{Habash2011,guinn2012language}\\
Pronoun Rate (PR)&Syntactic&Cognitive strength& \cite{Habash2011,guinn2012language}\\
Adjective Rate (AR)&Syntactic&Cognitive strength& \cite{Habash2011,guinn2012language}\\
Verbal Rate (VR)&Syntactic & Cognitive strength& \cite{Habash2011,guinn2012language}\\
Utterances&Lexical &the linguistic strength&\cite{orimaye2017predicting}\\
Function words&Lexical&---& \cite{orimaye2017predicting}\\
Word count&Lexical&---& \cite{orimaye2017predicting}\\
Character length&Lexical&---& \cite{orimaye2017predicting}\\
Total sentences&Lexical&---& \cite{orimaye2017predicting}\\
Unique words&Lexical&language processing& \cite{orimaye2017predicting}\\
Repetitions&Lexical&---& \cite{orimaye2017predicting}\\
Revisions&Lexical&---& \cite{orimaye2017predicting}\\
Morphemes&Lexical&---& \cite{orimaye2017predicting}\\
Trailing off indicator&Lexical&---& \cite{orimaye2017predicting}\\
Word replacement&Lexical&---& \cite{orimaye2017predicting}\\
Incomplete words&Lexical&---& \cite{orimaye2017predicting}\\
Filler words&Lexical&---& \cite{orimaye2017predicting}\\
Type token ration (TTR)&Semantic& Vocabulary Richness&\cite{Habash2011,guinn2012language}\\
Brunet's index (BI) &Semantic& Vocabulary Richness&\cite{Habash2011,guinn2012language}\\
Honore's statistics (HS)&Semantic& Vocabulary Richness&\cite{Habash2011,guinn2012language}\\
Fillers &Pragmatic& Cognitive Lapse&\cite{Habash2011,guinn2012language}\\
GoAhead utterances &Pragmatic&Cognitive functionality &\cite{Habash2011,guinn2012language}\\
Repetitions &Pragmatic&  Cognitive Lapse&\cite{Habash2011,guinn2012language}\\
Incomplete words &Pragmatic&Cognitive lapse& \cite{Habash2011,guinn2012language}\\
Syllables Per Minute &Pragmatic&Cognitive impairment & \cite{Habash2011,guinn2012language}\\
\hline
\end{tabular}
\label{tab1_L}
\end{table}

\begin{table}[t]
\caption{\footnotesize{List of Acoustic Features reported in the Literature within AD Domain.}}
\label{table:tab_acoustics_lit}
\setlength{\tabcolsep}{4pt}
\begin{tabular}{|p{180pt}|p{180pt}|p{80pt}|}
\hline
Type & Name & References \\
\hline
Cepstral Coefficients & Mean of MFCCs & \cite{Yancheva2015using, fraser2016linguistic} \\
 & Kurtosis of MFCCs & \cite{Yancheva2015using, fraser2016linguistic, hernandez2018computer} \\
  & Skewness of MFCCs & \cite{Yancheva2015using, fraser2016linguistic, hernandez2018computer} \\
Pauses and fillers & Total duration of pauses & \cite{Yancheva2015using, Roark2011spoken,konig2015automatic} \\
 & Mean duration of pauses & \cite{Yancheva2015using, Roark2011spoken, konig2015automatic, Ahmed2013connected} \\
   & Median duration of pauses & \cite{konig2015automatic, konig2015automatic} \\
  & SD of the duration of pauses & \cite{konig2015automatic} \\
 & Long and short pause counts & \cite{Yancheva2015using, Roark2011spoken} \\
 & Pause to word ratio & \cite{Yancheva2015using, fraser2016linguistic, Roark2011spoken, konig2015automatic, Ahmed2013connected} \\
  & Percentage of voiceless segments & \cite{Meilan2012acoustic}\\
 & Fillers (um, uh) & \cite{Yancheva2015using, guinn2012language, Roark2011spoken} \\
 
Pitch and Formants & Mean of F0, F1, F2, F3 & \cite{Yancheva2015using} \\

  & Variance of F0, F1, F2, F3 & \cite{Yancheva2015using} \\
  & Mean, SD, Max and Min of F0 & \cite{Meilan2012acoustic} \\

Aperiodicity & Jitter &  \cite{Yancheva2015using,Meilan2012acoustic} \\ 
 & Shimmer & \cite{Yancheva2015using, Meilan2012acoustic} \\ 
 & Recurrence rate & \cite{Yancheva2015using} \\
 & Recurrence period density entropy & \cite{Yancheva2015using} \\
 & Determinism & \cite{Yancheva2015using} \\ 
 & Length of diagonal structures & \cite{Yancheva2015using} \\
 & Laminarity & \cite{Yancheva2015using} \\
 
Temporal aspects of the speech & Total duration & \cite{Yancheva2015using, Meilan2012acoustic} \\
sample &Phonation time & \cite{ Meilan2012acoustic}\\
&Speech rate, syllable/s & \cite{ Meilan2012acoustic}\\
&Articulation rate, syllable/s & \cite{ Meilan2012acoustic}\\

Others & Zero-crossing rate & \cite{Yancheva2015using} \\
 & Autocorrelation & \cite{Yancheva2015using} \\
 & Linear prediction coefficients & \cite{Yancheva2015using} \\
 & Transitivity & \cite{Yancheva2015using} \\
\hline
\end{tabular}
\end{table}

%\
\subsection{ML Classifier}
\label{Machine Learning Classifier} 
ML algorithms 
such as \textit{k-Nearest Neighbor}~(k\-NN)~\cite{guinn2012language}, \textit{Support Vector Machine} (SVM), \textit{Decision Trees} (DT) ~\cite{guinn2012language, orimaye-etal-2014-learning} and \textit{Random Forest} (RF) classifiers, \textit{emotional learning-inspired ensemble classifier}~ \cite{pasrapoor2013emotional, Parsa-2020_Alzheimer} as well as \textit{Deep Learning}~(DL) architectures \cite{karlekar2018detecting} 
can analyze language produced by individuals (e.g., patients and healthy subjects) to distinguish healthy subjects from patients with dementia. 
In more details, the ML algorithms can be trained by linguistic features to identify language performance deficits in elderly individuals~\cite{mccullough2019language, guinn2012language, orimaye-etal-2014-learning}.
One of the earliest studies to develop such an ML algorithm
was proposed using the SVM to detect voice impairments in patients with AD~\cite{godino2005support}. 
In another work~\cite{orimaye2017predicting}, an SVM classifier was trained by language features extracted from the DB dataset and could achieve 80\% accuracy in predicting probable AD. In \cite{noorian2017importance}, an SVM classifier was trained on a dataset that combined DB with Talk2ME (i.e, encompass 167 patients with AD and 187 health controls), and achieved 70\% accuracy. Another excellent results obtained from employing an SVM classifier on a dataset that combined DB and CCC with 15 healthy controls and 26 patients with AD. They showed that the SVM can distinguish patients and healthy controls with 75\% accuracy \cite{konig2015automatic, guinn2012language}. In \cite{konig2015automatic, guinn2012language}, the authors showed that kNN can distinguish patients with MCI from healthy subjects with 63\% accuracy, they also showed that employing \textit{Bayesian Network} on the CCC dataset, we can achieve 66\% accuracy. 

%Machine learning algorithms have been widely used to analyze language produced by individuals to distinguish the ML community \cite{godino2005support,guinn2012language,orimaye-etal-2014-learning,asgari2017predicting,karlekar2018detecting}.
This section aims to provide an overview of various ML algorithms and datasets that have been used to identify language impairments associated with AD and MCI. Based on the overview, we believe that AI-powered LA tools can be considered as quick, accurate, cost-effective, user-friendly, reliable and valid tools to detect language impairment in the older adults. Therefore, in the next section, we describe our approach to develop such a tool. 

\section{AI-powered Language Assessment Tool}
Our approach to develop an AI-powered language assessment tool uses the same sequential steps that we described in the previous section. 
%1) Collecting language datasets; 2) Extracting linguistic and acoustic features; 
%3) Employing feature selection methods such as \textit{Variance Threshold}~(VT) or \textit{Minimal Redundancy Maximal Relevance Criterion}~(mRMR) to select informative features.
%4) Training various ML classifiers such as SVM, DTs, Extra Trees (ETs) by features and evaluating their performances. 
%5) Selecting an ML classifier with the best performance to be the basis of the AI-powered language assessment tool. 

%\subsection{Talk2Me}
\subsection{Language Dataset}
We have extracted audio and text datasets of patients ($N$=9) with various types of dementia\footnote{\footnotesize{patients have been diagnosed by physician from three hospitals in Toronto}} as well as healthy controls ($N$=13) from a database, named Talk2Me~\footnote{\footnotesize{each subject has signed a consent form that has been provided approved by the Research Ethics Board protocol 31127 of the University of Toronto}}. The Talk2Me database contains speech data recorded using a web or phone interface. 
In more details, textual and audio responses have been collected from participants using a variety of language tasks such as the PD and SR tasks. 
%Using the Talk2Me web interface and the phone network, we have collected various types of language datasets. In this study, we have focused on the audio and textual data obtained from two PD and SR Tasks. 

%\paragraph{Talk2Me}
%\paragraph{participants}
%\paragraph{Stimuli}
  
%\subsection{ML-based Language Assessment Tools}

\subsection{Feature Engineering}

\subsubsection{Linguistic Features}
\label{linguistic_features}
We extract different linguistic features (e.g., the lexical diversity) from textual data using the \textit{Natural Language Toolkit}~\cite{Loper02nltk:the}. The linguistic features of this paper can be divided into three categories: 1) Lexical features (e.g., lexical richness); 2) Syntactic features (e.g., \textit{Part-of-Speech} (POS)); and 3) Semantic features.
\paragraph{\textbf{Lexical Features:}}
Since dementia can influence the lexical richness of patients' language, different studies have proposed different types of lexical features as markers of language impairment in patients with AD/MCI. For example, 
in \cite{orimaye2017predicting}, utterances, word count, character length, total sentences, unique words, repetitions, revisions, morphemes, incomplete words, filler words, trailing off indicator, and 
word replacement extracted as lexical features. However, in our study, we have 
extracted multiple features such as \textit{Brunet\textsc{\char13} s index}~(BI)
(see Equation \ref{eq:Lex_1}) and \textit{Honor\textsc{\char13}s Statistic} (HS) with Equation \ref{eq:Lex_2} \cite{malvern2004lexical} to measure the lexical richness. In Equations \ref{eq:Lex_1} and \ref{eq:Lex_2}, $w$ and $u$ are the total number of word tokens and the total number of unique word types, respectively. There are five readability scores namely the \textit{Flesch-Kincaid} ($F_K$) (see Equation \ref{eq:Lex_3}), the \textit{Flesch Reading-Ease}~(FRES) Test (see Equation \ref{eq:Lex_4})~\cite{kincaid1975derivation}, to test the readability of the transcripts. Here, $s$ and $SYL$ indicate the total number of sentences and the total number of syllables, respectively.
\begin{equation}
\label{eq:Lex_1}
BI=w^{(u^{-0.165})}
\end{equation} 
\begin{equation}
\label{eq:Lex_2}
HS=\frac{100\log w}{1-\frac{w}{u}}
\end{equation} 
\begin{equation}
\label{eq:Lex_3}
F_K=0.39 \left ( \frac{w}{s} \right ) + 11.8 \left ( \frac{SYL}{w} \right ) - 15.59
\end{equation}
\begin{equation}
\label{eq:Lex_4}
FRES=206.835 - 1.015 \left( \frac{w}{s} \right) - 84.6 \left( \frac{SYL}{w} \right)
\end{equation}
\paragraph{\textbf{Syntactic Features:}}
We have also extracted syntactic features such as POS ratios: 1) third pronouns (3rd-pron-pers) to proper nouns (prop); 2) first pronouns (1st-pron-pers) to pronouns (1st-pron-pers) \footnote{\footnotesize{People with dementia may use first person singular pronouns than physicians perhaps as a way of focusing attention on their perspective \cite{sakai2011linguistic}
}}; 3) nouns to verbs; and 4) \textit{subordinate} to \textit{coordinate}~\cite{komeili2019talk2me} to calculate syntactical error in speech, which is indicative of \textit{frontotemporal dementia}~\cite{peelle2007syntactic}, and propositional and content density equations \ref{eq:Syn_1} and \ref{eq:Syn_2} to quantify the syntax complexity. Here, $NN$, $VB$, $JJ$, $RB$, $IN$, and $CC$ are the number of nouns, verbs, adjectives, adverbs, prepositions, and conjunctions respectively.
\begin{equation}
  \label{eq:Syn_1}
    density_{p}=\frac{VB+JJ+RB+IN+CC}{N}
\end{equation}
\begin{equation}
 \label{eq:Syn_2}
    density_{c}=\frac{NN+VB+JJ+RB}{N}
\end{equation}
\paragraph{\textbf{Semantic-based Features:}}
Patients with dementia cannot easily retrieve semantic knowledge, reflecting a semantic decline in their language \cite{konig2015automatic}. 
To develop a tool that can detect semantic decline and also incoherent speech, tangentiality \cite{Parsa-NN2020}, we suggest training ML algorithms using extracted semantic-based features, which are referred as incoherent  and tangential metrics in this paper.  
The incoherence metrics are extracted by calculating themilarity (Equation~\ref{eq:sem_1}) between sentence embeddings: ${v}_{s_{j}}$. Various sentence embeddings such as \textit{Simple Average} (SA)\footnote{SA provides sentence embedding by averaging generated word embeddings from text files.}(see Equation~\ref{eq:sem_2}), or \textit{Smooth Inverse Frequency} (SIF) embeddings \footnote{SIF provides sentence embedding by calculating the weighted average of word embeddings and removing their first principal component}~\cite{arora2016simple} (see Equation~\ref{eq:sem_3})
and \textit{term frequency-Inverse Document Frequency}~(tf-IDF) (see Equation~\ref{eq:sem_4}).
We have also calculated a tangential metric employing \textit{Latent Dirichlet Allocation}~\cite{blei2003latent, landauer1998introduction, https://doi.org/10.48550/arxiv.2009.13602} (see Equation \ref{eq:T_1}). Using the tangential metric, we can measure tangentiality \cite{Parsa-NN2020} in speech of patients with dementia. We measured tangential speech using Equation \ref{eq:T_1}. Here, $N_{topic}$ is the optimal number of topics for a corpus made of interview of subjects \cite{https://doi.org/10.1002/alz.12278}.
\begin{equation}
\label{eq:sem_1}
Similarity_{\textbf{SA}} ({v}_{s_{i}},{v}_{s_{j}})= {{v}_{s_{i}} \cdot {v}_{s_{j}} \over \|{v}_{s_{i}}\| \|{v}_{s_{j}}\|}
\end{equation}

\begin{equation}
\label{eq:sem_2}
Similarity_{\textbf{SIF}} ({v}_{s_{i}},{v}_{s_{j}})= 1-{{v}_{s_{i}} \cdot {v}_{s_{j}} \over \|{v}_{s_{i}}\| \|{v}_{s_{j}}\|}
\end{equation}
\begin{equation} 
\label{eq:sem_3}
Incoherence_{\textbf{SA}}=\min_{i}{\max_{j}{ Similarity_{\textbf{SA}}  ({v}_{s_{i}},{v}_{s_{j}})\over{abs(i-j)+1} 
}}
\end{equation} 

\begin{equation} 
\label{eq:sem_4}
Incoherence_{\textbf{SIF}}= \min_{i}{\sum_{j}{Similarity_{\textbf{SIF}} ({v}_{s_{i}},{v}_{s_{j}})\over{abs(i-j)+1} 
}}
\end{equation} 
\begin{equation} 
\label{eq:sem_5}
Incoherence_{\textbf{TFIDF}}=
\min_{i}{\sum_{j}{Similarity_{\textbf{TFIDF}} ({v}_{s_{i}},{v}_{s_{j}})\over{abs(i-j)+1} 
}}
\end{equation} 

\begin{equation}
\label{eq:T_1}
  Tangentiality=1-\frac{N_{topic}}{\sum_{j}{N_{topic}}}
\end{equation} 
\subsubsection{Acoustic Features}
\label{acoustic_features} 
We have extracted the acoustic features using the \textit{COre Variable Feature Extraction Feature Extractor}~(COVFEFE) tool~\cite{komeili2019talk2me}. We have considered 37 acoustic features and their \textit{Mean}, \textit{Standard Deviation} (std), \textit{Skewness} (skew) (lack of symmetry of a data distribution) and \textit{Kurtosis} (kurt) (measure of peakedness around the mean of a data distribution) which resulted in a total of 148 features. We have also included the deltas of these 148 features. Therefore, our feature selection methods have considered 296 features in total. For example, we have considered mean, std, skew and kurt of an MFCC feature (described later) and its deltas. Thus from a single acoustic feature, we have extracted 8 additional features. We have followed the same procedure to extract all 296 features. We have divided our features in 3 groups: 1) Spectral Features, 2) Phonation and  Voice  Quality  Features, and 3) Speech  Features. Table~\ref{table:tab_acoustics_feat} shows the list of features that we have considered in the research. In this section, we only describe the features that are identified as meaningful by our feature selection methods.  

\begin{table}[t]
\caption{\footnotesize{List of Acoustic Features that are Considered in this Research}}
\label{table:tab_acoustics_feat}
\setlength{\tabcolsep}{4pt}
\begin{tabular}{|p{200pt}|p{100pt}|p{100pt}|p{40pt}|}
\hline
Type & Name & Functional & \# of Features \\
\hline
Spectral Features & MFCCs 0 - 14  & mean, kurt, skew, std & 60 \\
                  & $\Delta$ MFCCs 0 - 14  & mean, kurt, skew, std & 60 \\
                  & log Mel freq 0 - 7    & mean, kurt, skew, std & 32 \\
                  & $\Delta$ log Mel freq 0 - 14  & mean, kurt, skew, std & 32 \\
                  & LSP freq 0 - 7  & mean, kurt, skew, std  & 32\\
                  & $\Delta$ LSP freq 0 - 7  & mean, kurt, skew, std & 32\\
\hline
Phonation and Voice  & F0  & mean, kurt, skew, std & 4 \\
Quality Features & $\Delta$ F0  & mean, kurt, skew, std & 4 \\
                  &  Jitter local           & mean, kurt, skew, std & 4 \\
                  &  $\Delta$ Jitter local  & mean, kurt, skew, std & 4\\
                  &  Jitter DDP  & mean, kurt, skew, std & 4 \\
                  &  $\Delta$ Jitter DDP  & mean, kurt, skew, std & 4 \\
                  &  Shimmer  & mean, kurt, skew, std & 4 \\
                  &  $\Delta$ Shimmer  & mean, kurt, skew, std & 4 \\
                  &  Loudness  & mean, kurt, skew, std & 4 \\
                  &  $\Delta$ Loudness  & mean, kurt, skew, std & 4 \\
\hline                 
Speech Features  & Voicing prob.  & mean, kurt, skew, std & 4\\
                  & $\Delta$ Voicing prob. & mean, kurt, skew, std & 4\\
                  
\hline

\end{tabular}
\end{table}

\paragraph{\textbf{Spectral Features:}}
We have considered the features derived from the \textit{Mel Frequency Cepstrum} (MFC) and the \textit{Line Spectral Pairs} (LSPs) to develop our ML classifiers.
MFC uses the Mel scale to represent short-term power spectrum of a sound. \textit{Mel Frequency Cepstral Coefficients} (MFCCs) represent energy variations between frequency bands of a speech signal and are effectively used for speech recognition and speaker verification. MFCCs aim at accurately representing the phonemes articulated by speech organs (tongue, lips, jaws, etc.). Delta MFCCs are the trajectories of the MFCCs over time. The logarithm of Mel filter banks  are calculated as an intermediate step of computing MFCCs and we have considered the \textit{Log Mel Frequency Bands} and the \textit{Delta Log Mel Frequency Bands} as spectral features. Previous research identified the  MFCCs as one of the most relevant acoustic features to distinguish patients with different types of dementia~\cite{Yancheva2015using, fraser2016linguistic, hernandez2018computer}. Our analysis also confirm this claim (see Tables~\ref{tab_com_feat_task_all} and \ref{tab_com_feat_media_all}). 

LSPs are strongly related to underlying speech features and are thus useful in speech coding~\cite{McLoughlin2008Line}. They are correlated to unvoiced speech, pause and silence which are reportedly effective in identifying linguistic impairments~\cite{mcloughlin1999lsp}. The delta of LSPs represents the change of LSPs over time. Our feature selection methods confirm the importance of LSPs and their deltas (see Tables~\ref{tab_com_feat_task_all} and \ref{tab_com_feat_media_all}).  

\textbf{Phonation and Voice Quality Features:}
This feature group includes \textit{Fundamental Frequency} (F0), \textit{Shimmer}, \textit{Jitter}, \textit{Loudness}, and the deltas of these features. The F0 feature is defined as the rate of oscillation of the vocal folds~\cite{de2002yin}. F0 is nearly periodic in speech of the healthy people but less periodic in patients~\cite{tsanas2011nonlinear}. Jitter describes frequency instability and shimmer is a measure of amplitude fluctuations. Loudness affects the amplitude of vibrations and it is correlated to the emotional states of the speaker~\cite{yanushevskaya2013voice}. Previous studies reported that phonation and voice quality features are correlated with MCI and AD~\cite{meilan2014speech, lopez2013automatic}, and our findings also support these claims (see Tables~\ref{tab_com_feat_task_all} and \ref{tab_com_feat_media_all}).

\textbf{Speech Features:}
We have considered the \textit{Voicing Probability} and the delta of voicing probability as relevant acoustic features. A voicing probability shows the percentage of unvoiced and voiced energy in a speech signal. A delta voicing probability indicates the rate of change over time. Our feature selection methods identified that mean, std and kurt of both features are discriminative features to identify older adults living with dementia~(see Table~\ref{tab_com_feat_media_all}). 

\subsection{ML Classifiers}
The ML classifier are employed to analyze linguistic and acoustic features,  which have been extracted from individuals' language. For this paper,  
we have trained different classifiers such as DT, \textit{Extra Tree} (ET), kNN, SVM using a set of extracted linguistic and acoustic features and evaluated their performances. We have selected a classifier with higher performance to be the basis of the LA tool. 

\section{Results}
We have employed different ML algorithms and trained them on 
various language features extracted from subjects' speeches during the PD and SR tasks and these speeches have been collected using phone-based and web-based interfaces. We have compared the performances of the ML classifiers to verify the language task's impact and recording media on accuracy and reliability of our suggested approach. 
Note that, we have trained the classifiers separately with linguistic and acoustic features, and therefore, in the following parts, we compare the performance of the classifiers developed with these two groups of features.

%The results verify 1) a language task that can be quick and accurate to distinguish language impairment; 2) a set of language markers can accurately train ML algorithms that robustly, precisely, and efficiently (high specificity and sensitivity) distinct patients with language impairments from healthy subjects. 

%and 3) to develop ML-based assessment tools. three experiments that we have designed to evaluate the efficiency of using traditional ML tools to develop ML-based assessment tools. 

%In each experiment, we employ ML tools (Table \ref{tabel:ML_tools}) on a new dataset (see Table (\ref{DATA})) and evaluate the results to find the ML tool with highest accuracy, precision, specificity and sensitivity. 
\subsection{Language Tasks}
This subsection investigates the impact of the two language tasks (see Table \ref{tab:data}) the PD and SR tasks on the performance of classifiers at the subject level. These two tasks assess different cognitive characteristics of patients with dementia, and thus it is worth investigating and comparing the effectiveness of these two language tasks.   
%In this section, we compare the ML algorithms developed by using the stimuli collected during the PD and SR language assessment tasks. The PD and SR tasks are well-explored in the literature for language-based assessment. These two tasks assess different cognitive characteristics of the patients, and thus it is worth investigating and comparing the effectiveness of these two language tasks in dementia assessment. 

\begin{table}[!h]
\caption{Statistics about our textual datasets}  
\label{stat}
\setlength{\tabcolsep}{6pt}
\begin{tabular}{|p{130pt}|p{70pt}|p{70pt}|p{70pt}|p{70pt}|p{70pt}}
\hline
DATA & Ave Sentence & Std Sentence & Ave Word& Std Word \\
\hline
The PD Task& 9.0 & 4.4 & 153.5 & 97.92\\
\hdashline
The SR Task & 6.79 & 4.00 & 57.07 & 26.91 \\
\hdashline
Recording Media (Phone) & 3.5 & 4.66& 74.0& 44.90\\
\hdashline
Recording Media (Web) & 2.27& 1.25& 65.59& 31.11\\
\hline
\end{tabular}
\end{table}

\subsubsection{The PD Task}
\label{pd_task}
\begin{figure}
    \centering
    {\includegraphics[scale=.3]{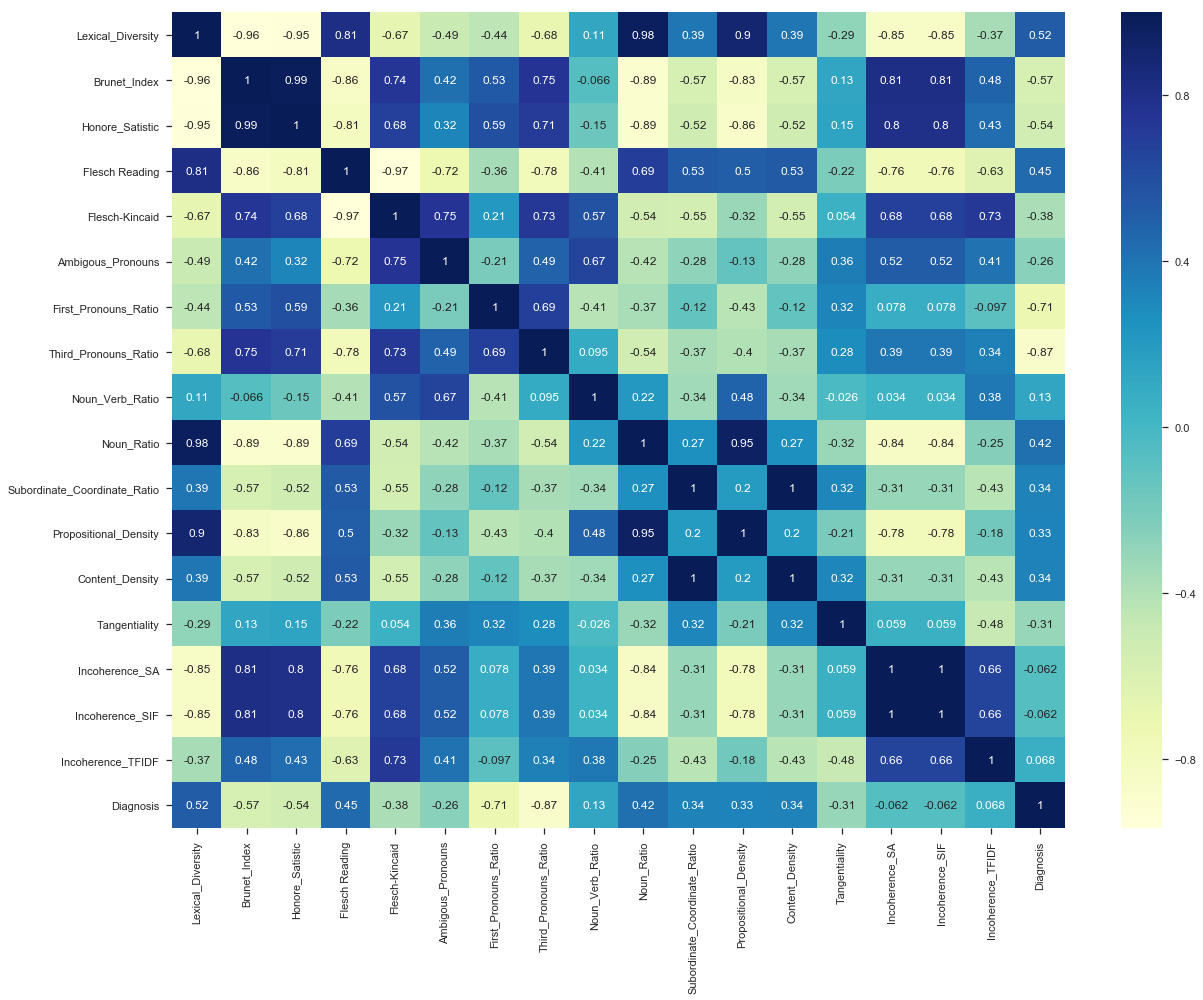}}
    \caption{Correlation heat-map between 17 linguistic features. }
    %Training ETs with these features, it can classify two classes with an accuracy of 63\% ); however DTs can achieve 73\% to classify the classes.}
    \label{fig:Feature_Correlation_AD_PD}
\end{figure}

We study the efficacy of linguistic (see Figure~\ref{fig:Feature_Correlation_AD_PD}) and acoustic features (see Table~\ref{tab_com_feat_task_all}), which have been extracted from the speech of the subjects without dementia ($N$=3) and subjects with dementia ($N$=5) during completing the PD task on the performance of our proposed approach. 
%which is used to develop the AI-powered assessment tool for dementia. To do so, we train different ML algorithms with various sets of features and analyze the results in the following sections.   

\begin{figure}
\begin{subfigure}{.45\textwidth}
  \centering
  \includegraphics[width=1\linewidth]{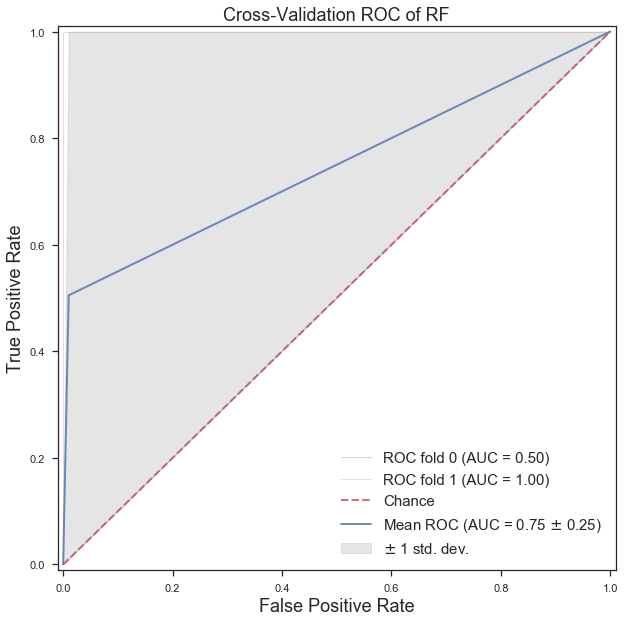}
  \caption{ROC curves of RF trained by 17 linguistic features}
\end{subfigure}%
\begin{subfigure}{.45\textwidth}
  \centering
  \includegraphics[width=1\linewidth]{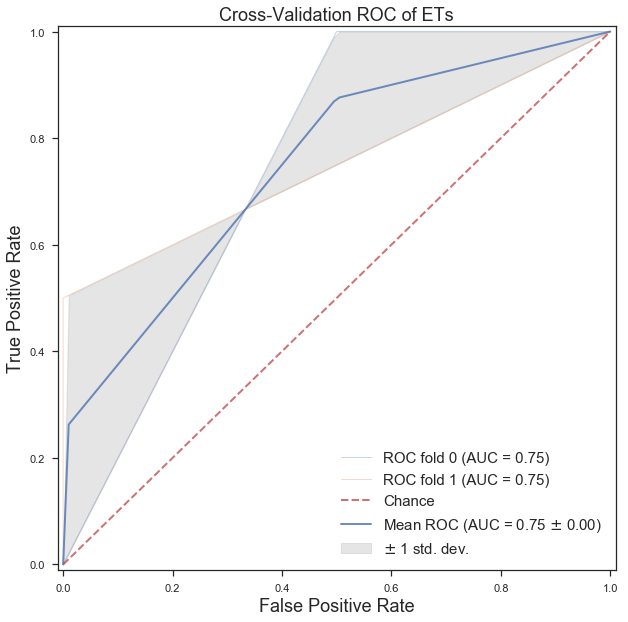}
  \caption{ROC curves of ETs by lexical features}
\end{subfigure}%
\caption{ROC curves of RF and ETs trained by different sets of linguistic features}
\label{fig:ROC_1}
\end{figure}

\begin{figure}
\begin{subfigure}{.45\textwidth}
  \centering
  \includegraphics[width=1\linewidth]{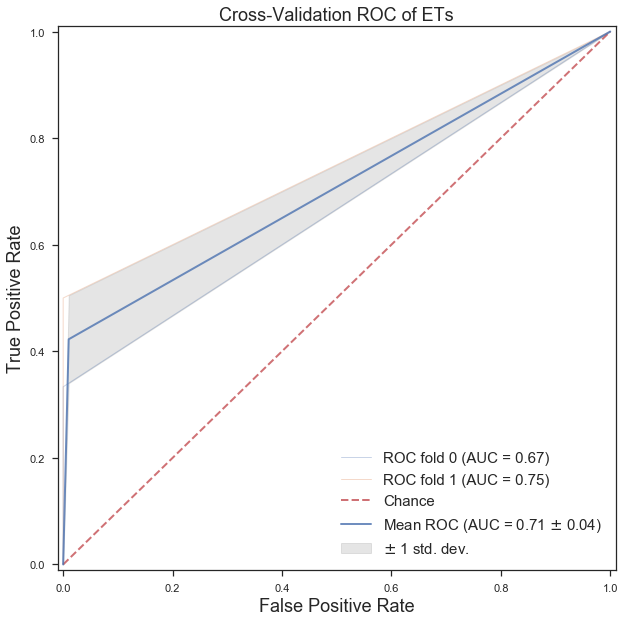}
  \caption{ROC curves of ETs trained by syntactic features }
\end{subfigure}
\begin{subfigure}{.45\textwidth}
  \centering
  \includegraphics[width=1\linewidth]{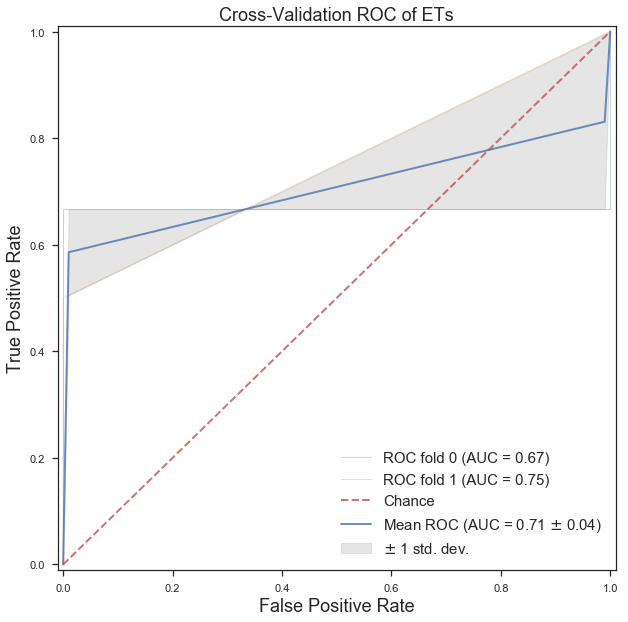}
  \caption{ROC curves of ET trained by semantic features}
\end{subfigure}
\caption{ROC curves of ETs trained by syntactic and semantic features}
\label{fig:ROC_2}
\end{figure}

%that correctly detect language impairment of older adults using the PD task. 
%For this purpose, we train the ML algorithms with linguistic (see Figure~\ref{fig:Feature_Correlation_AD}) and acoustic features (see Table~\ref{tab_com_feat_all}) obtained from the participants' speech during the PD task.
%We are particularly interested to evaluate the importance  of different types of features and the performance of different ML algorithms in assessing language impairment associated to patients with dementia.
\paragraph{Classifiers with Linguistic Features}
We trained various ML algorithms using lexical, semantic and syntactic features, which have been extracted from the textual datasets obtained from speech datasets. Figure \ref{fig:ROC_1}.(b) shows that if we train the ET algorithm with a set of lexical features, we can achieve 
more accurate classification results than other ML algorithms. Training ML algorithms with the set of lexical, semantic and syntactic features decreases the accuracy of classifiers (see Figure \ref{fig:ROC_1}.(a)). By training the ML classifiers using 8 Syntactic features, we observed the ET algorithm could classify the classes with an accuracy of 63.0\% (+/-7\%). By training various ML classifiers using 4 semantic features, we observed ET provide more accurate results than others and could classify the classes with an accuracy of 63.0\% (+/-7\%) (see Figure \ref{fig:ROC_2}.(b)).

\begin{figure}[h!]
    \centering
    {\includegraphics[scale=.55]{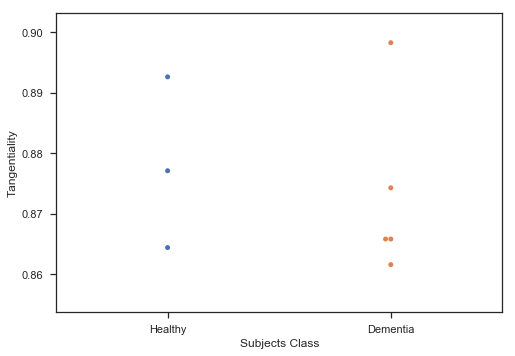}}
    \caption{A comparison between the tangentiality measure for subjects with and without dementia.}
    \label{fig:Tangentiality_PD}
\end{figure}

\begin{figure}
  \centering
  \includegraphics[scale=.55]{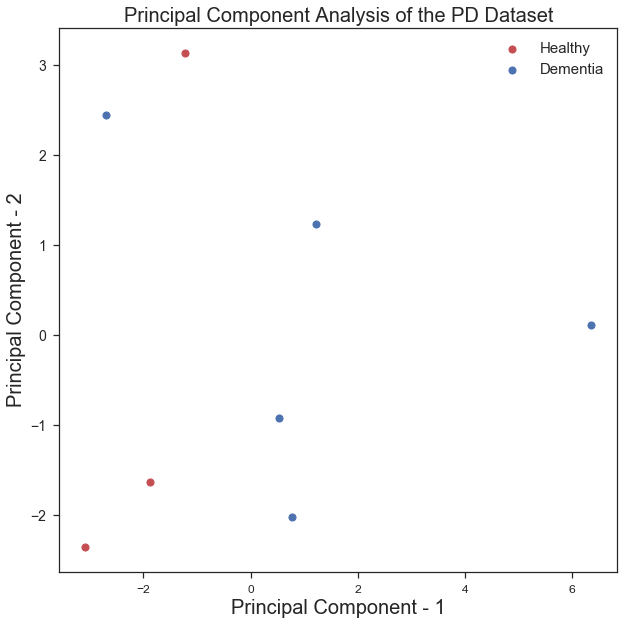}
  \caption{It shows that subjects with dementia and healthy controls cannot be linearly separated using 2 principle components.}
\label{fig:PCA}
\end{figure}

%Figures \ref{fig:Corr_AD}, \ref{fig:Corr_AD_SY} and \ref{fig:Corr_AD_Sem} present the correlation between different set of features and two classes of patients and healthy subjects. 
Training ML algorithms with 3 principle components (see Figure \ref{fig:PCA}) extracted from 17 features, we observed that the SVM algorithm with the linear kernel could classify with 63.0\% (+/- 7\%) accuracy. 
Furthermore, among lexical features, two Flesch-Kincaid ( CV\footnote{coefficient of variation}=23.17\%, p\_value=0.25) and Flesch-Reading-Ease (CV=15.15\%, p\_value=0.35) can provide better discrimination between these two groups of subjects, while the number of the third pronouns (the effect size equals to 1.319) and the first pronouns (the effect size equals to 2.198) among subjects without dementia has higher value than subjects with dementia. Thus, these two syntactic features can be considered as markers to detect subjects with MCI. Another interesting result is that measuring tangentiality (see
Figure \ref{fig:Tangentiality_PD}) (with the effect size of 0.020) in speech can provide a better understanding to determine subjects with dementia from healthy subjects. 

\begin{figure}
  \centering
  \includegraphics[scale=.55]{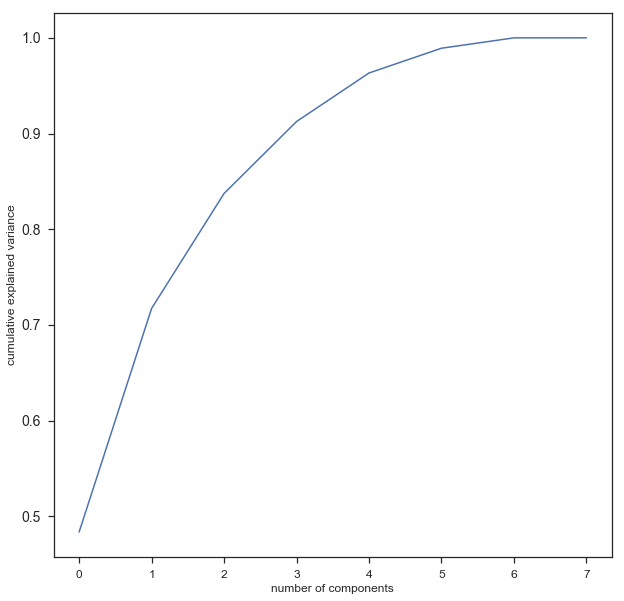}
  \caption{It presents the values of cumulative explained variance for different number of principle components.}
\label{fig:PCA_CEV}
\end{figure}
%can provide they can classify patients and healthy subjects with 63\% accuracy. 

%Table \ref{tab_LF_results_all} presents the classification results obtained from applying multiple ML algorithms trained  
%If we train ML algorithms mentioned in the Table \ref{tab_AF_results_all} using 15 features, most of them (such as RF, SVM, ETs) can classify patients and healthy subjects with 63\% accuracy. Using 5 lexical features (see Figure \ref{Corr_AD}), ETs can achieve 73\% (+/- 0.13), SVMs can provide the same results, other classifier can classify patients and healthy control less accurate than being trained by all features. 
%Using 8 syntactic features (see Figure \ref{fig:Corr_AD_SY}) to train classifiers SVMs can achieve 63\% accuracy. Using 4 semantic features (see Figure \ref{fig:Corr_AD_Sem}) RF and SVMs can determine patients from healthy subjects with 73\% (+/- 0.13) and 63(+/- 0.0)\% accuracy, respectively. 

%\textcolor{red}{Section~\ref{linguistic_features} lists the linguistic features that we consider in this research. We rank the top 8 features using Shpiro ranking and use these feature to classify the healthy control and AD/MCI. We use default scikit-learn configuration for all algorithms and evaluate the performance using 3-fold cross-validation........Parse please explain how you get the results in Table \ref{tab_LF_results_all}......}   

\paragraph{Classifiers with Acoustic Features} 
Table~\ref{tab_AF_results_all} presents the classification results obtained by applying ML tools on the extracted features from the audio files. The ML classifiers are trained using the spectral (e.g., MFCC, LSP), speech (e.g., voicing probability), phonation (e.g., F0) and voice quality (e.g., jitter, shimmer) features as described in Section~\ref{acoustic_features}. We rank all these features using \textit{Analysis of Variance} (ANOVA), RF and ~\textit{Minimal Redundancy Maximal Relevance }~(mRMR) methods and use the top 8 features identified by each of these methods to train the ML classifiers. Table~\ref{tab_com_feat_task_all} shows the top common acoustic features ranked by the above mentioned feature selection methods. We found that scikit-learn's \cite{pedregosa2011scikit} default configurations work fine for the considered ML classifiers. Therefore, we use the default configurations for all classifiers. The F1 micro scores in Table~\ref{tab_AF_results_all} are obtained using the 3-fold cross-validation method. Our results show that the tree-based classifiers, e.g., RF, ET, and DT outperform the others. 

%Table~\ref{tab_AF_results_all} presents the classification results obtained by applying ML tools on the features extracted from the audio files. We use the same dataset as we used in the the previous section. The ML classifiers are developed using the spectral (e.g., MFCC, LSP), speech (e.g., voicing probability), phonation (e.g., F0) and voice quality (e.g., jitter, shimmer) features as described in Section~\ref{acoustic_features}. We rank all these features using ANOVA, RF and mRMR methods and use the top 8 features identified by each of these methods to develop the ML classifiers. Table~\ref{tab_com_feat_task_all} shows the top common acoustic features ranked by these 3 feature selection methods. We found that scikit-learn's \cite{pedregosa2011scikit} default configurations work fine for the considered ML classifiers. Therefore, we use the default configurations for all classifiers. The F1 micro scores in Table~\ref{tab_AF_results_all} are obtained using the 3-fold cross-validation method. Our results show that the tree based classifiers, e.g., RF, ET and DT provide better performance than others. 

\begin{table}[h!]
\caption{Common acoustic features obtained by applying ANOVA, RF and mRMR feature selection methods on the recorded audio files of the PD and SR tasks}
\label{tab_com_feat_task_all}
\setlength{\tabcolsep}{4pt}
\begin{tabular}{|p{220pt}|p{220pt}|}
\hline
PD Task & SR Task \\
\hline
MFCC 13 (mean) & MFCC 12,13 (skew) \\
MFCC 12,13 (kurt) & $\Delta$ MFCC 3,4,6,13 (mean) \\
MFCC 10,13 (skew) & $\Delta$ MFCC 4 (skew) \\
$\Delta$ MFCC 2,11 (mean) & $\Delta$ LSP freq 7 (mean) \\
$\Delta$ MFCC 2,3,6,7 (kurt) & $\Delta$ LSP freq  3,6 (kurt) \\
$\Delta$ MFCC 6,11,13 (skew) & Loudness (kurt, skew) \\
$\Delta$ LSP freq 3,5 (mean)&$\Delta$ Loudness (kurt, skew)\\
$\Delta$ LSP freq 2,6 (kurt) & F0 (kurt)\\
$\Delta$ LSP freq  1 (skew) & $\Delta$ F0 (mean)  \\

\hline
\end{tabular}
\end{table}

\begin{table}[h!]
\caption{F1 (micro) scores obtained by applying ML algorithms on linguistic features}
\label{tab_LF_results_all}
\setlength{\tabcolsep}{6pt}
\begin{tabular}{|p{40pt}|p{70pt}|p{70pt}|p{70pt}|p{70pt}|p{70pt}|}
\hline
Features & Algorithms & PD Task & SR Task & Web & Phone \\
\hline
Lexical  & DT & 0.63 (+/- 0.07) & \textbf{0.71 (+/- 0.00)} &0.42 (+/- 0.17) &  0.92 (+/- 0.17) \\ 
        & ET & \textbf{0.73 (+/- 0.13)} & 0.57 (+/- 0.57) &0.83 (+/- 0.00) &   0.92 (+/- 0.17) \\  
        & kNN & 0.52 (+/- 0.29) & 0.42 (+/- 0.00) & 0.45 (+/- 0.00)&  0.45 (+/- 0.00)\\ 
        & LDA & 0.63 (+/- 0.07)& 0.63 (+/- 0.07) & 0.75 (+/- 0.17)& 0.92 (+/- 0.17) \\ 
        & R\_SVM & 0.63 (+/- 0.07) & \textbf{0.71 (+/- 0.00)} &0.83 (+/- 0.00) & 0.83 (+/- 0.00) \\  
        & L\_SVM & 0.63 (+/- 0.07)& \textbf{0.71 (+/- 0.00)} &0.83 (+/- 0.00) &  1.00 (+/- 0.00)  \\
        & LR & 0.63 (+/- 0.07)& 0.63 (+/- 0.07) & 0.83 (+/- 0.00)& 1.00 (+/- 0.00)\\ 
        & RF & 0.47 (+/- 0.27) & \textbf{0.71 (+/- 0.00)} &0.83 (+/- 0.00) &0.92 (+/- 0.17)\\   
\hdashline

Syntactic  & DT &0.73 (+/- 0.13) & 0.57 (+/- 0.00) &0.83 (+/- 0.00) &  0.83 (+/- 0.00) \\ 
        & ET & 0.80 (+/- 0.40) & 0.64 (+/- 0.14)  &0.83 (+/- 0.00) &  0.83 (+/- 0.00) \\  
        & kNN & 0.69 (+/- 0.63) & 0.53 (+/- 0.23) &0.45 (+/- 0.00) &  0.45 (+/- 0.00)\\ 
        & LDA & 0.37 (+/- 0.07) & 0.50 (+/- 0.43) &  0.75 (+/- 0.17)& 0.75 (+/- 0.50) \\ 
        & R\_SVM & 0.63 (+/- 0.07)& \textbf{0.71 (+/- 0.00)} & 0.83 (+/- 0.00) & 0.83 (+/- 0.00) \\  
        & L\_SVM & 0.63 (+/- 0.07) & \textbf{0.71 (+/- 0.00)}  &0.83 (+/- 0.00) & 0.67 (+/- 0.33)
 \\
        & LR & 0.80 (+/- 0.40) & \textbf{0.71 (+/- 0.00)} & 0.83 (+/- 0.00) & 0.75 (+/- 0.17) \\ 
        & RF & 0.47 (+/- 0.27) & 0.57 (+/- 0.00) &0.75 (+/- 0.17) & 0.92 (+/- 0.17) \\ 
\hdashline

Semantic& DT & 0.53 (+/- 0.27) & 0.64 (+/- 0.14) & 0.83 (+/- 0.00) & 0.83 (+/- 0.33) \\ 
        & ET & 0.57 (+/- 0.47) & 0.71 (+/- 0.29) &0.83 (+/- 0.00) & 0.83 (+/- 0.00) \\  
        & kNN & 0.69 (+/- 0.63) & 0.53 (+/- 0.23) & 0.45 (+/- 0.00) &0.45 (+/- 0.00)  \\ 
        & LDA & 0.63 (+/- 0.07) & \textbf{0.71 (+/- 0.00)} & 0.83 (+/- 0.00) & 0.58 (+/- 0.50) \\ 
        & R\_SVM & 0.63 (+/- 0.07) & \textbf{0.71 (+/- 0.00)} & 0.83 (+/- 0.00) & 0.83 (+/- 0.00) \\  
        & L\_SVM & 0.63 (+/- 0.07) & 0.71 (+/- 0.00) & 0.83 (+/- 0.00) &0.83 (+/- 0.00)   \\
        & LR & 0.63 (+/- 0.07) & 0.50 (+/- 0.43) & 0.83 (+/- 0.00) & 0.83 (+/- 0.00) \\ 
        & RF & \textbf{0.73 (+/- 0.13)} & 0.57 (+/- 0.00) &0.83 (+/- 0.00) & 0.83 (+/- 0.00) \\ 
\hline
All     & DT & \textbf{0.73 (+/- 0.13)} & 0.64 (+/- 0.14)  & 0.75 (+/- 0.17) &1.00 (+/- 0.00  \\ 
        & ET & 0.63 (+/- 0.07) & \textbf{0.79 (+/- 0.14)}  &  0.83 (+/- 0.00) &  0.75 (+/- 0.50)\\  
        & kNN & 0.52 (+/- 0.29) & 0.39 (+/- 0.05) &  0.45 (+/- 0.00) & 0.45 (+/- 0.00)  \\ 
        & LDA & 0.63 (+/- 0.07) & 0.64 (+/- 0.14) & 0.75 (+/- 0.17) & 0.75 (+/- 0.50 \\ 
        & R\_SVM & 0.63 (+/- 0.07) & 0.71 (+/- 0.00) & 0.83 (+/- 0.00) &0.83 (+/- 0.00)  \\  
        & L\_SVM & 0.63 (+/- 0.07) & 0.64 (+/- 0.14) & 0.75 (+/- 0.17) &  1.00 (+/- 0.00)\\
        & LR & 0.70 (+/- 0.60) & 0.64 (+/- 0.14)  &  0.83 (+/- 0.00 &0.75 (+/- 0.17)  \\ 
        & RF & 0.63 (+/- 0.07) & 0.71 (+/- 0.00) &  0.83 (+/- 0.00) & 1.00 (+/- 0.00) \\ 
         
\hline
\end{tabular}
\end{table}

\enlargethispage{10pt}

\begin{table}[h!]
\caption{F1 (micro) scores obtained by applying ML algorithms on acoustic features}
\label{tab_AF_results_all}
\setlength{\tabcolsep}{6pt}
\begin{tabular}{|p{40pt}|p{70pt}|p{70pt}|p{70pt}|p{70pt}|p{70pt}|}
\hline
Features & Algorithms & PD Task & SR Task & Web & Phone\\
\hline
ANOVA  & DT &  0.83 (+/- 0.24) & 0.50 (+/- 0.24) & \textbf{0.89 (+/- 0.16)} & 0.81 (+/- 0.02)\\ 
        & ET & 0.98 (+/- 0.03) & \textbf{0.86 (+/- 0.09)} & 0.83 (+/- 0.24) & 0.93 (+/- 0.09)\\  
        & kNN & 0.83 (+/- 0.24) & 0.78 (+/- 0.02) & \textbf{0.89 (+/- 0.16)} & 0.93 (+/- 0.09)\\ 
        & LDA & 0.89 (+/- 0.16) & 0.70 (+/- 0.14) & \textbf{0.89 (+/- 0.16)} & \textbf{1.00 (+/- 0.00)} \\ 
        & R\_SVM & 0.72 (+/- 0.21) & 0.78 (+/- 0.02) & \textbf{0.89 (+/- 0.16)} & 0.76 (+/- 0.06)\\  
        & L\_SVM & 0.83 (+/- 0.24) & 0.70 (+/- 0.14) & 0.72 (+/- 0.21) & \textbf{1.00 (+/- 0.00)}\\
        & LR & 0.83 (+/- 0.24) & 0.78 (+/- 0.02) & 0.72 (+/- 0.21) & 0.93 (+/- 0.09)\\ 
        & RF & \textbf{0.99 (+/- 0.02)} & 0.83 (+/- 0.06) & 0.83 (+/- 0.24) & 0.93 (+/- 0.09)\\   
\hline
RF      & DT & 0.72 (+/- 0.21) & 0.57 (+/- 0.17)  & \textbf{1.00 (+/- 0.00)} & 0.87 (+/- 0.09)\\ 
        & ET & 0.99 (+/- 0.02) & 0.80 (+/- 0.04)  & \textbf{1.00 (+/- 0.00)} & 0.99 (+/- 0.02)\\  
        & kNN & 0.89 (+/- 0.16) & 0.78 (+/- 0.02) & 0.89 (+/- 0.16) & 0.93 (+/- 0.09)\\ 
        & LDA & \textbf{1.00 (+/- 0.00)} & 0.57 (+/- 0.17) & 0.89 (+/- 0.16) & 0.93 (+/- 0.09)\\ 
        & R\_SVM & 0.61 (+/- 0.08) & 0.78 (+/- 0.02) & 0.61 (+/- 0.08) & 0.76 (+/- 0.06)\\  
        & L\_SVM & 0.89 (+/- 0.16) & 0.78 (+/- 0.02) & 0.72 (+/- 0.21) & \textbf{1.00 (+/- 0.00)} \\
        & LR & 0.89 (+/- 0.16) & \textbf{0.87 (+/- 0.09)} & 0.72 (+/- 0.21) & 0.93 (+/- 0.09)\\ 
        & RF & \textbf{1.00 (+/- 0.00)} & 0.78 (+/- 0.02) & 0.90 (+/- 0.14) & \textbf{1.00 (+/- 0.00)}\\ 
\hdashline
mRMR    & DT & \textbf{1.00 (+/- 0.00)} & 0.70 (+/- 0.14) & 0.83 (+/- 0.24) &  0.87 (+/- 0.09)\\ 
        & ET & \textbf{1.00 (+/- 0.00)} & \textbf{0.81 (+/- 0.05)}  & 0.97 (+/- 0.05) & \textbf{1.00 (+/- 0.00)}\\  
        & kNN & 0.50 (+/- 0.14) & 0.78 (+/- 0.02) & \textbf{1.00 (+/- 0.00)}& 0.81 (+/- 0.16)\\ 
        & LDA & \textbf{1.00 (+/- 0.00)} & 0.77 (+/- 0.21) &\textbf{1.00 (+/- 0.00)} & \textbf{1.00 (+/- 0.00)}\\ 
        & R\_SVM & 0.61 (+/- 0.08) & 0.78 (+/- 0.02) & 0.72 (+/- 0.21) & 0.76 (+/- 0.06)\\  
        & L\_SVM & 0.78 (+/- 0.31) & 0.50 (+/- 0.08)  & \textbf{1.00 (+/- 0.00)} & 0.87 (+/- 0.19)\\
        & LR & 0.78 (+/- 0.31) & 0.78 (+/- 0.02) & \textbf{1.00 (+/- 0.00)} & 0.87 (+/- 0.19)\\ 
        & RF & 0.99 (+/- 0.02) & 0.78 (+/- 0.02) & 0.88 (+/- 0.16) & \textbf{1.00 (+/- 0.00)}\\ 
\hdashline
Common  & DT & \textbf{1.00 (+/- 0.00)} & 0.52 (+/- 0.37)  & \textbf{1.00 (+/- 0.00)} & 0.80 (+/- 0.28)\\ 
        & ET & \textbf{1.00 (+/- 0.00)} & \textbf{0.84 (+/- 0.11)} & 0.74 (+/- 0.21) & 0.94 (+/- 0.08)\\  
        & kNN & 0.83 (+/- 0.24) & 0.70 (+/- 0.14) & 0.89 (+/- 0.16) & \textbf{1.00 (+/- 0.00)}\\ 
        & LDA & 0.78 (+/- 0.16) & 0.80 (+/- 0.16) &0.89 (+/- 0.16) & 0.87 (+/- 0.19)\\ 
        & R\_SVM & 0.72 (+/- 0.21) & 0.78 (+/- 0.02) & 0.78 (+/- 0.16)& 0.76 (+/- 0.06)\\  
        & L\_SVM & 0.83 (+/- 0.24) & 0.77 (+/- 0.21)  &0.72 (+/- 0.21) & \textbf{1.00 (+/- 0.00)}\\
        & LR & 0.83 (+/- 0.24) & 0.70 (+/- 0.14) &0.72 (+/- 0.21) & \textbf{1.00 (+/- 0.00)}\\ 
        & RF & 0.98 (+/- 0.02) & 0.78 (+/- 0.02) & 0.81 (+/- 0.20) & 0.95 (+/- 0.07)\\ 
         
\hline
\end{tabular}
\end{table}

\subsubsection{The SR Task}
\label{sr_task}
\begin{figure}
\begin{subfigure}{.45\textwidth}
  \centering
    \includegraphics[width=1\linewidth]{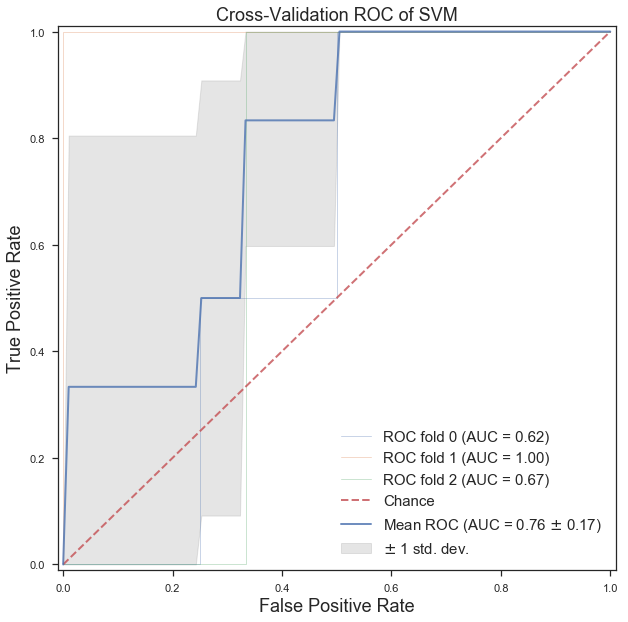}
  \caption{ROC curves of SVM trained by 17 features}
\end{subfigure}%
\begin{subfigure}{.45\textwidth}
  \centering
  \includegraphics[width=1\linewidth]{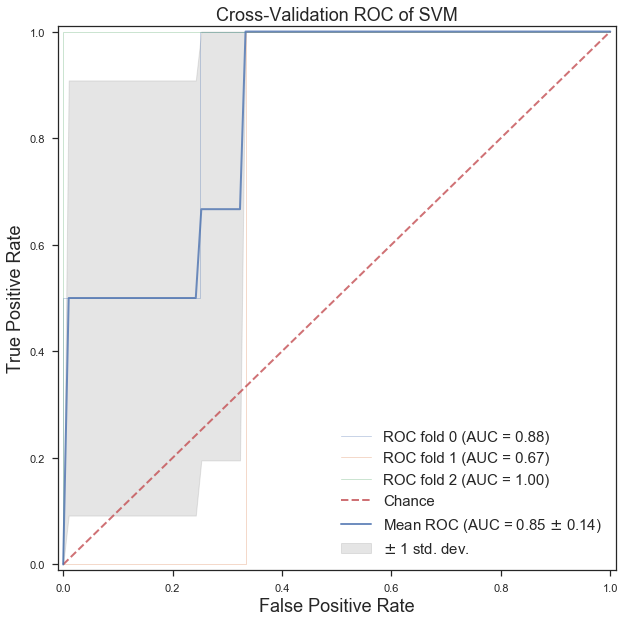}
  \caption{ROC curves of SVM trained by lexical features }
\end{subfigure}
\caption{ROC curves of SVM trained by all and lexical features}
\label{fig:ROC_SR_1}
\end{figure}

\begin{figure}[h!]
    \centering
    {\includegraphics[scale=.55]{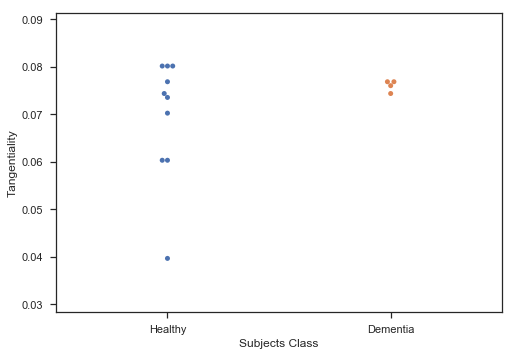}}
    \caption{A comparison between the tangentiality measure for subjects with dementia and healthy subjects.}
    \label{fig:Tangentiality_SR}
\end{figure}

\begin{figure}
  \centering
  \includegraphics[scale=.55]{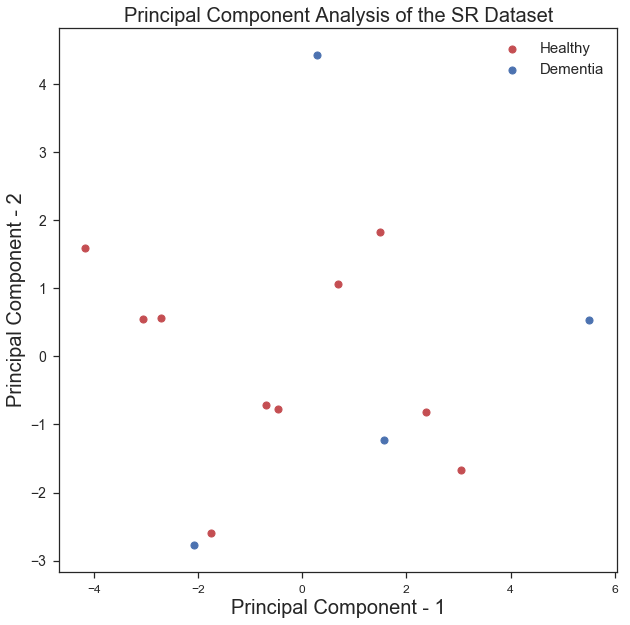}
  \caption{It shows that subjects with dementia and healthy controls cannot be linearly separated using 2 principle components.}
\label{fig:PCA_SR_1}
\end{figure}

\begin{figure}
  \centering
  \includegraphics[scale=.55]{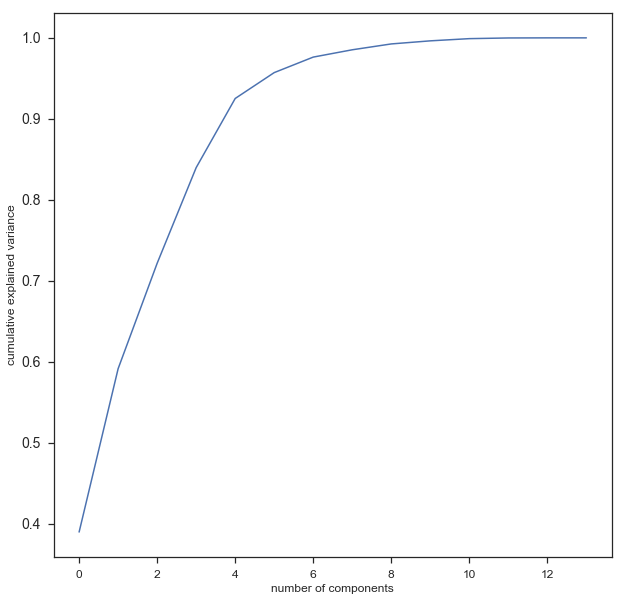}
  \caption{It presents the values of cumulative explained variance for different number of principle components.}
\label{fig:PCA_CEV_SR}
\end{figure}

This section presents the results obtained by training different ML classifier using linguistic and acoustic features extracted from language data produced subjects without dementia ($N$=10) and subjects with dementia ($N$=4) during the SR task.
\paragraph{Classifiers with Linguistic Features}
This section examines the efficiency of using different linguistic features to train ML classifiers. 
%To do so, we have extracted different sets of different linguistic features. 
Using 5 lexical features to train classifiers, the SVM (with the \textit{Radial Basis Function}~(RBF) kernel and $C$=0.01) and RF ($n\_estimators$=2 and $max\_depth$=2), can classify subjects with dementia and healthy subjects accurately with 71\% accuracy. We can get the same results using
8 syntactic features to train the SVM (with the RBF kernel and $C$=0.01) and RF($n\_estimators$=2 and $max\_depth$=2). 
If we train the classifiers (3-fold Cross-Validation) with 17 lexical, semantic, and syntactic features, the SVM (with the RBF kernel and $C$=0.01) can classify subjects with dementia and healthy adults accurately with 72\% accuracy (see Figure \ref{fig:ROC_SR_1}.(a)). It is interesting that measuring tangentiality (see Figure \ref{fig:Tangentiality_SR}) 
in speech can provide a better understanding to determine subjects with dementia from healthy subjects. 
Training ML algorithms with 3 principle components (see Figures \ref{fig:PCA_SR_1} and \ref{fig:PCA_CEV_SR}) extracted from 17 features, we observed that the SVM algorithm with the RBF kernel could classify with 71\% accuracy. 

%Training a 2D CNN with filter and kernel size of 4 and 2 with the textual data sets, we can achieve 71.43\% accuracy to distinguish health subjects from patients with AD.

\begin{figure}

\begin{subfigure}{.45\textwidth}
  \centering
  \includegraphics[width=1\linewidth]{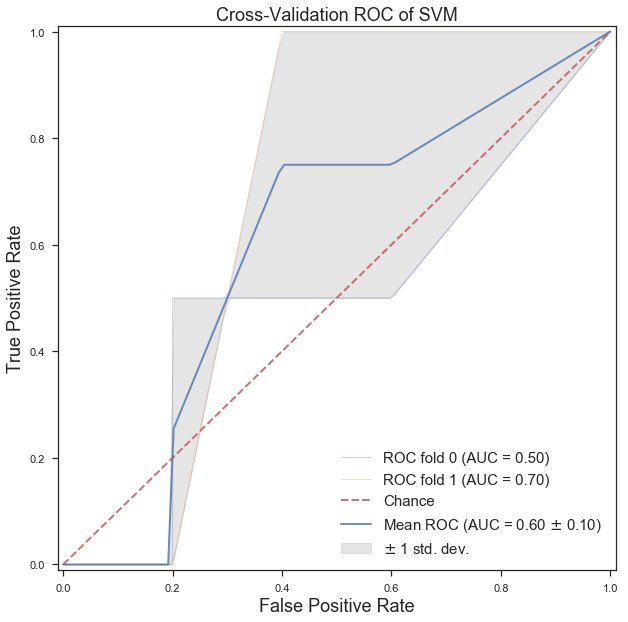}
  \caption{ROC curves of SVM trained by syntactic features}
\end{subfigure}
\begin{subfigure}{.45\textwidth}
  \centering
  \includegraphics[width=1\linewidth]{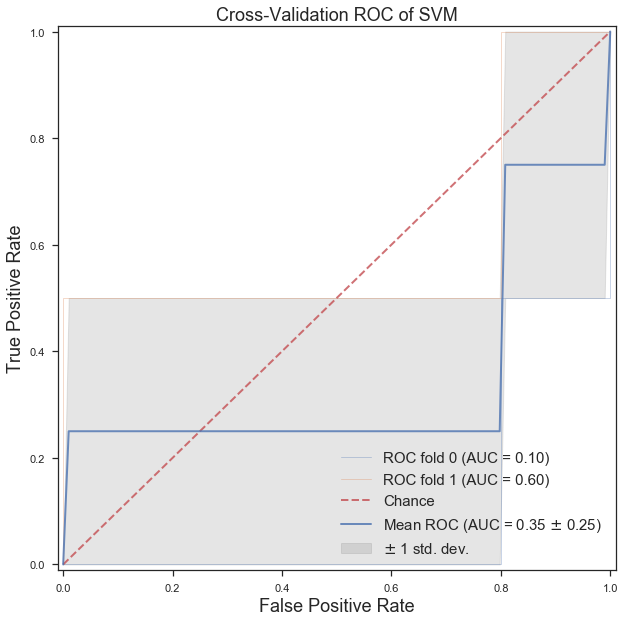}
  \caption{ROC curves of SVM trained by 4 semantic features}
\end{subfigure}
\caption{ROC curves of SVM trained by syntactic and semantic features}
\label{fig:ROC_SR_2}
\end{figure}
\begin{figure}[h!]
    \centering
    {\includegraphics[scale=.3]{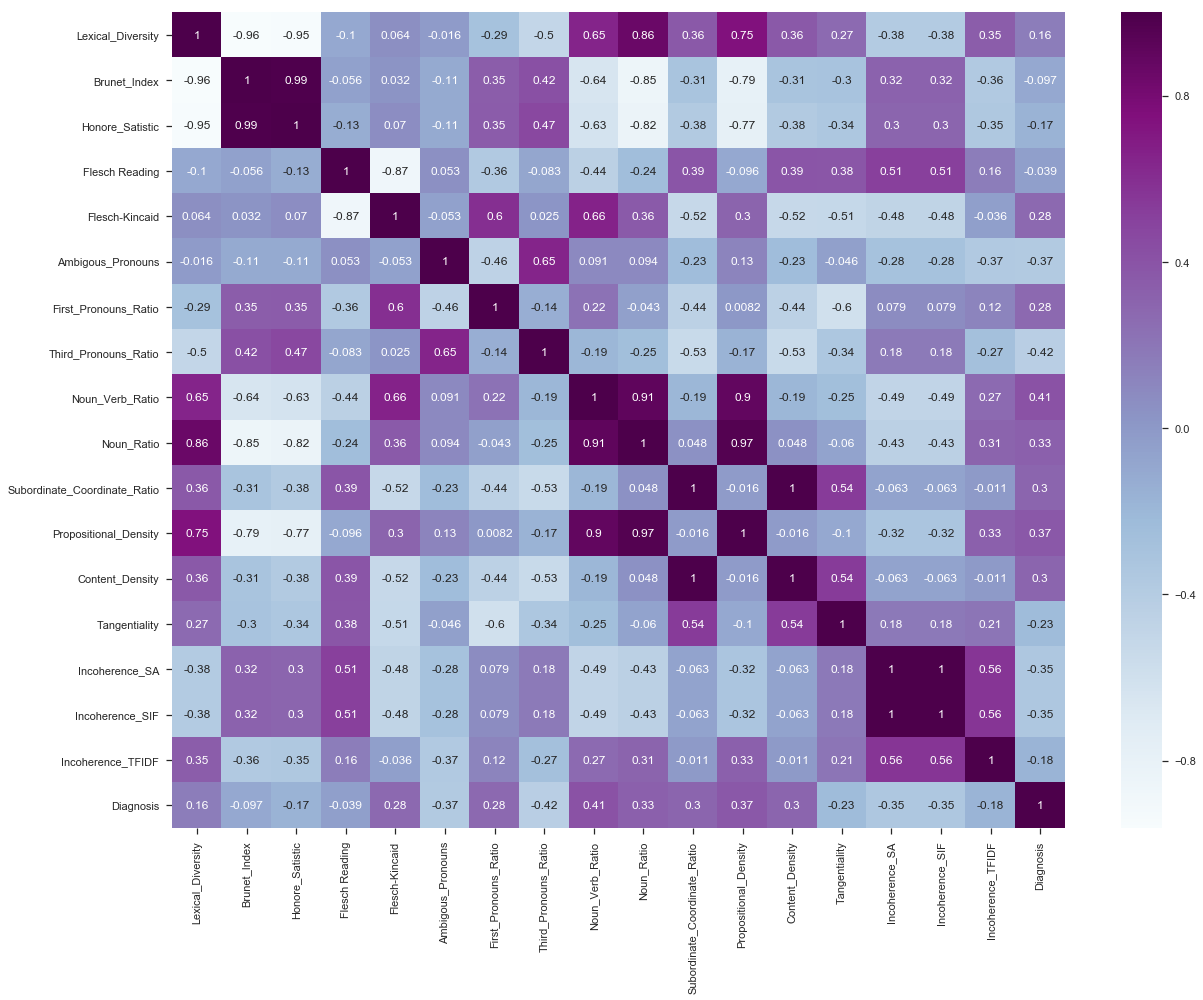}}
    \caption{The correlation between different features}
    \label{fig:Feature_corrrelation_SR}
\end{figure}

\paragraph{Classifiers with Acoustic Features}
In this section, we present the classification results obtained by applying ML algorithms on the features extracted from the audio files. We use audio data collected from subjects without dementia  ($N=10$) and subjects with dementia ($N=4$). We follow the same approach that we use in Section \ref{pd_task} to rank the acoustic features and use the top 15 features to develop the classifiers. Table~\ref{tab_com_feat_task_all} shows the top common acoustic features ranked by ANOVA, RF and mRMR feature selection methods. We use scikit-learn's default configurations for all classifiers of this sub-section. The F1 micro scores in Table~\ref{tab_AF_results_all} are obtained using the 3-fold cross-validation method.
Our results show that the ET classifiers outperform others.

%\begin{figure}
%    \centering
%    \begin{minipage}{0.45\textwidth}
%        \centering
%        \includegraphics[width=0.9\textwidth]{bmc_template/images/boxplot_F1_task.png} % first figure itself
%        \caption{first figure}
%    \end{minipage}\hfill
%    \begin{minipage}{0.45\textwidth}
%        \centering
%        \includegraphics[width=0.9\textwidth]{bmc_template/images/boxplot_F1_task.png} % second figure itself
%        \caption{second figure}
%    \end{minipage}
%\end{figure}

\subsubsection{Comparison between the PD and the SR Tasks}

\begin{figure}
\begin{subfigure}{.5\textwidth}
  \centering
  \includegraphics[width=1\linewidth]{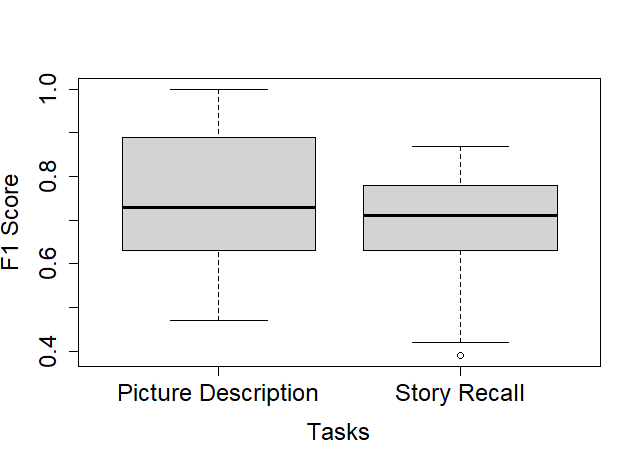}
  \caption{Picture Description vs Story Recall Task}\label{fig:boxplot_tasks}
\end{subfigure}%
\begin{subfigure}{.5\textwidth}
  \centering
  \includegraphics[width=1\linewidth]{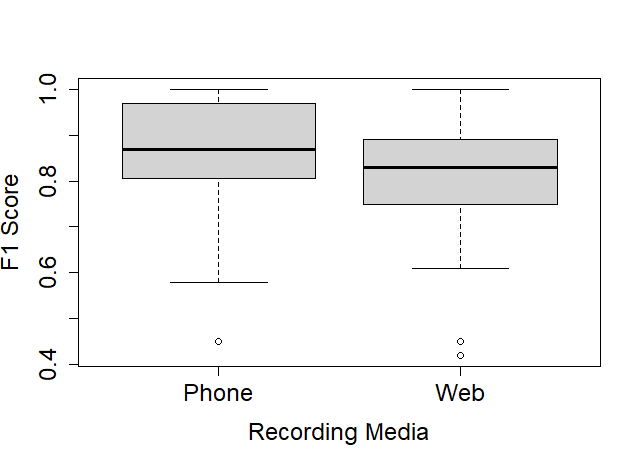}
  \caption{Web vs Phone-based Recordings}\label{fig:boxplot_media}
\end{subfigure}
\begin{subfigure}{.5\textwidth}
  \centering
  \includegraphics[width=1\linewidth]{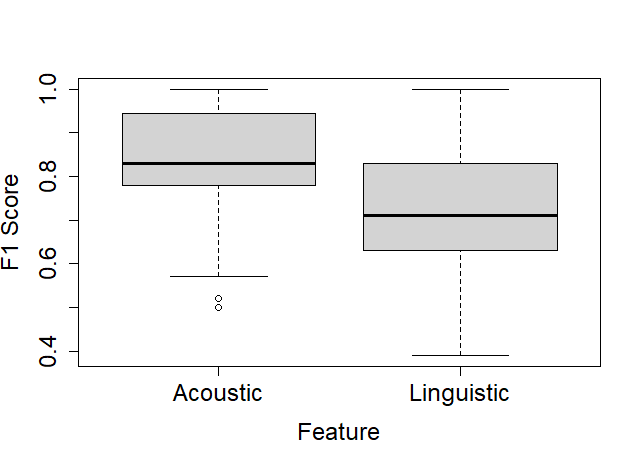}
  \caption{Acoustic vs Linguistic Features}\label{fig:boxplot_features}
\end{subfigure}
\caption{Boxplots showing the F1 scores obtained from different classifiers: \newline a) distributions of the F1 scores in the picture description and story recall tasks, b) distributions of the F1 scores for web and phone-based recordings, and  \newline c) distributions of the F1 scores in the linguistic and acoustic features. \newline These boxplots show that the picture description task, phone-based recordings and acoustic features provide better performance than the story recall task, web-based recorings and linguistic features.}\label{fig:boxplot_all}
\end{figure}

As mentioned earlier, we have also evaluated the impact of language tasks on the accuracy of our approach in detecting AD and MCI. We have used audio recordings and transcribed textual datasets to extract linguistic and acoustic features from PD and SR tasks. Our datasets are imbalanced and therefore micro F1 scores are more appropriate to report the performance of the ML classifiers. To asses the efficiency of PD and SR tasks, we have calculated a range of F1 scores using different feature sets and classifiers as shown in Tables~\ref{tab_LF_results_all} and ~\ref{tab_AF_results_all}. We have used lexical, syntactic, semantic and combination of all these 3 feature groups as linguistic features. For acoustic features, we have used ANOVA, RF and mRMR feature selection methods. We have also used the common features in these 3 feature selection methods as another set of acoustic features. Finally, we have apply DT, ET, Linear SVM, RBF SVM, \textit{Linear Discriminant Analysis}~(LDA), \textit{Logistic Regression}~(LR), kNN and RF algorithms to compute the F1 scores. Figure \ref{fig:boxplot_all}(a) shows the distributions of F1 scores for PD and SR tasks. A one-way ANOVA test performed on the F1 scores of the PD and SR tasks shows that the means are significantly different (F(1,126) = 8.27, \textit{p} = 0.005). A Tukey's post-hoc test shows that the mean F1 scores of the PD task are higher than the SR task (\textit{p} = 0.005), i.e., the ML classifiers developed by using PD stimuli perform better than the SR stumuli. 

%Shapiro-Wilk normality tests show that F1 scores for PD (p-value = 0.0006) and SR (p-value = 0.002) tasks are not normally distributed (p-value $<$ 0.05). Therefore, we cannot apply parametric tests (e.g., ANOVA) to compare means of the F1 scores of these 2 tasks. A Wilcoxon test shows that mean F1 score of the PD task are higher than the SR task (p-value = 0.02), i.e., the ML algorithms developed by using PD stimuli perform better than the SR stumuli.

%\begin{figure}[t!]
%    \centering
%    {\includegraphics[scale=.50]{bmc_template/images/Boxplot_PD_vs_SR.png}}
%    \caption{Boxplot showing the F1 scores of the classifiers that are developed \newline  using the picture description and story recall tasks}
%    \label{fig:boxplot_tasks}
%\end{figure}

%\begin{figure}[h!]
%    \centering
%    {\includegraphics[scale=.50]{bmc_template/images/Boxplot_Phone_vs_Web.png}}
%    \caption{Boxplot showing the F1 scores of the classifiers that are developed \newline using phone-based and web-based recordings}
%    \label{fig:boxplot_media}
%\end{figure}

%\begin{figure}[h!]
%    \centering
%    {\includegraphics[scale=.50]{bmc_template/images/Boxplot_Ling_vs_Acous.png}}
%    \caption{Boxplot showing the F1 scores of the classifiers that are developed  \newline using the linguistic and acoustic features}
%    \label{fig:boxplot_features}
%\end{figure}

%\setcounter{subfigure}{1}

\subsection{Recording Media}
We were also interested in figuring out the direct effect of using web-interface or phone-interface on the quality of recorded language data that indirectly affect on the accuracy of AI-powered LA Tools. 

\paragraph{Classifiers with Linguistic Features}
We have trained various ML classifiers using linguistic features extracted from recorded language data (12 samples, 10 samples related to subjects without dementia and 2 samples related to subjects with dementia) that were collected using the phone-interface and the web-interface. Table \ref{tab_LF_results_all} shows that the results obtained from the web-interface data is more accurate than the results obtained from the phone-interface data. Using 5 lexical features, the SVM (with the linear kernel) classifier and the LR can classify samples with 99.9\% accuracy. However, using 8 syntactic features, we drop all ML classifiers' performance, including the SVM (with the linear kernel); thus, the SVM can determine subjects with dementia with 83\% accuracy. Similarly, if we use 4 semantic features to train classifiers, they can provide better performance than using 8 syntactic features.

%Figure \ref{fig:EXP3_Two_D_Features} describes a two-dimensional ranking of 15 features using the Pearson correlation score to detect a co-linear relationship between pairs of features) to train ML classifier correctly.

\paragraph{Classifiers with Acoustic Features}
We have developed the classifiers using the acoustic features extracted from the the audio files. We used 16 phone-based recordings from 3 healthy adults and 1  dementia patient (each participant attended 4 sessions). Similarly, we have considered 8 web-based recordings from subjects with dementia ($N=3$) and subjects with dementia ($N=5$) (only 1 session each). We have followed the same approach to rank the acoustic futures as we described before for the acoustic features. Table~\ref{tab_com_feat_media_all} shows the common features ranked by ANOVA, RF and mRMR methods. We use the top 15 features to train the classifiers. Table~\ref{tab_AF_results_all} shows the F1 scores obtained from the DT, ET, Linear SVM, RBF SVM, LDA, LR, kNN and RF algorithms. We have used scikit-learn’s default configurations and the 3-fold cross-validation method to calculate the F1 scores. We found that DT perform better for web-based recordings and linear the SVM showed better performance for phone-based recordings.

\begin{table}[h!]
\caption{Common acoustic features obtained by applying ANOVA, RF and mRMR feature selection methods on phone and web-based recordings}
\label{tab_com_feat_media_all}
\setlength{\tabcolsep}{4pt}
\begin{tabular}{|p{220pt}|p{220pt}|}
\hline
Web & Phone \\
\hline
MFCC 5,11,12 (mean)  & MFCC 6, 9 (std)\\
$\Delta$ MFCC 11, 13 (mean) & MFCC 3 (skew)\\
$\Delta$ MFCC 0,3,6,9,10 (skew) & MFCC 3, 5 (kurt)\\
$\Delta$ log Mel freq 0,5,6 (skew) & $\Delta$ MFCC 0 (std)\\
Voicing prob. (kurt, std) & LSP freq 7 (mean)\\
$\Delta$ Voicing prob. (kurt, mean, std) & LSP freq 2, 3, 4 (skew) \\
LSP freq 0 (kurt) & LSP freq 1 (kurt)\\
F0 (skew) & $\Delta$ LSP freq 3 (mean)\\
Jitter local (kurt, skew) & $\Delta$ LSP freq 5 (skew)\\
$\Delta$ Jitter local (kurt) & log Mel freq 2 (skew)\\
$\Delta$ Jitter DDP (kurt) & $\Delta$ log Mel freq 1, 2, 3 (std)\\
$\Delta$ Shimmer local (kurt) & Voicing prob. (kurt, std)\\
                             & Loudness (kurt)\\
\hline

\end{tabular}
\end{table}

%\begin{table}[H]
%\caption{\footnotesize{Comparison between various ML algorithms to distinguish healthy subjects and patients from analyzing extracted linguistic features from language data produced during the PD task}.}
%\label{LF_PD}
%\setlength{\tabcolsep}{6pt}
%\begin{tabular}{|p{60pt}|p{50pt}|p{58pt}|p{58pt}|}
%\hline
%Task  & Features& ML algorithm&
%                    Model's Accuracy  \\
%\hline
   
%   PD&Syntactic&ETs& 0.90 (+/- 0.20)\\
%   SR&Lexical &R\_SVM& 0.71 (+/- 0.00) \\
%   SR (Phone)&Lexical& DTs &1.00 (+/- 0.00)\\
%   SR (WEb)& Syntactic&ETs &0.85 (+/- 0.22) \\
   
%\hline
%\end{tabular}
%\label{LF}
%\end{table}
\subsubsection{Comparison between the phone-based and the web-based interfaces}  
We have performed a one-way ANOVA test on the F1 scores of the phone and web-based recordings as shown in Tables ~\ref{tab_LF_results_all} and ~\ref{tab_AF_results_all}. Our analysis shows that the means of these 2 groups are significantly different (F(1,126) = 4.26, \textit{p} = 0.04). Figure~\ref{fig:boxplot_all}(b) shows the distributions of F1 scores of these 2 groups.
A Tukey's post-hoc test shows that the mean F1 scores of the classifiers developed by the extracted features from the phone-based recordings are higher than web-based recordings (\textit{p} = 0.04), i.e., the ML classifiers trained with the phone-based recordings perform better than the web-based  recordings. 
%As evident from the reported results, the ML-based tools provide more accurate results when data is collected using the phone-interface. 

\subsection{Linguistic and Acoustic Features}
Tables~\ref{tab_LF_results_all} and ~\ref{tab_AF_results_all} show the results obtained by using different linguistic features and acoustic features. We consider all F1 scores (total 256) to compare the performance between the classifiers trained with linguistic and acoustic features. Figure~\ref{fig:boxplot_all}(c) shows the distributions of F1 scores of these 2 groups. A one-way ANOVA test performed on the F1 scores shows that the means are significantly different (F(1, 256) = 62.43, \textit{p} $\approx$ 0). A Tukey's test for post-hoc analysis shows that the mean F1 scores of the classifiers trained with the acoustic features are higher than the classifiers trained with the linguistic features (\textit{p} = 0). That is, the ML classifiers trained with the acoustic features perform better than the classifiers trained with the linguistic features.   

\section{Discussion}
This paper proposed an approach to develop an AI system to recognize different dimensions of language impairments in subjects with dementia. This section discusses what is the limitation of such a development and how we can develop a valid, reliable, fair, and explainable AI-based LA tool. 
\subsection{Data Limitation} 
\begin{figure}[h!]
    \centering
    {\includegraphics[scale=.7]{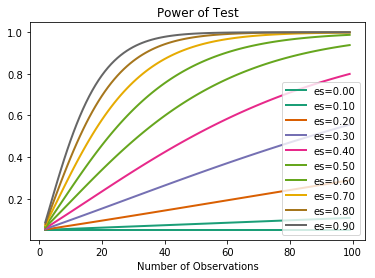}}
    \caption{A description of using a power analysis to estimate the minimum sample size is required for achieving a desired effect size; It shows the impact of different effect sizes (es) and various sizes of the data sample on the statistical power}
    \label{fig:Sample_Size}
\end{figure}

No doubt having a lot of data samples, ML algorithms, which are cores of our AI-powered LA tool
can learn better~\cite{domingos2012few} to map linguistic features to the group of subjects (i.e., with dementia or without dementia). 
In other words, determining the optimal sample size for developing an efficient AI-powered LA tool assures an adequate power to detect statistical significance \cite{suresh2012sample}. However, for our problem, collecting language data from too many subjects is expensive and needs a lot of time. 
Thus, even it is necessary to estimate what is the sufficient size of samples for achieving acceptable classification results and then start to develop an AI-powered assessment tool, but our results have shown that we could achieve good performance even with using the language data of less than 10 subjects (see Figure \ref{fig:Sample_Size}). 
\subsection{End-to-end Learning Approach}
To develop an AI-powered language assessment tool, we can use an end-to-end learning approach. Table~\ref{DL} presents the obtained results employing \textit{Convolutional Neural Networks} (ConvNet/CNN) on classifying the textual datasets. It shows that we can obtain approximately same results using CNN to classify our dataset. It might be a good idea to employ deep learning algorithm if we have sufficient samples. 
\begin{table}[h!]
\caption{Results from CNN}
\label{DL}
\setlength{\tabcolsep}{3pt}
\begin{tabular}{|p{120pt}|p{280pt}|p{50pt}|}
\hline
Experiment& Model Structure&
                    Accuracy\\
\hline
  PD task& 2-channel CNN, Filter Size= 4, Kernel Size=2& 66.7\%\\
   \hdashline
  SR task& 2-channel CNN, Filter Size= 4, Kernel Size=2& 71.4\%\\
   \hdashline
 Recording Media (Web) & 2-channel CNN, Filter Size= 4, Kernel Size=2& 80.0\%\\
  \hdashline
  Recording Media (PHONE) & 2-channel CNN, Filter Size= 4, Kernel Size=2& 40.0\%
\\
\hline
\end{tabular}
\end{table}

\subsection{Generalization - Selecting Meaningful Features}

One of the problems we have faced with the acoustic features is that when we have applied ANOVA, RF, and mRMR feature selection methods on different datasets (i.e., obtained from various recording media or language tasks), each time we have received different sets of features (see Tables~\ref{tab_com_feat_task_all} and~\ref{tab_com_feat_media_all}). Therefore, we are interested in combining all features so that we get almost consistent performance with all datasets. For this purpose, we have used~\textit{ Principal Component Analysis}~(PCA) to combine a group of features, as shown in Table~\ref{tab_pca_feat}. We have considered MFCCs (0 to 14th order coefficients), the deltas of these MFCCs, and the deltas of LSP frequency bands (0 to 7) in the PCA because these groups of features appear more frequently in our rankings (see Tables~\ref{tab_com_feat_task_all} and~\ref{tab_com_feat_media_all}). We found that the first two \textit{Principal Components}~(PCs) can retain, on average 75\% of the variance, and hence we have considered only the first 2 PCs to train the classifiers. Table~\ref{tab_pca_rslt} shows how these PCA features perform with 4 different sets of data. Our results show that we achieved almost consistent performance with the tree-based classifies ranging from 78\% to 93\% F1 scores with these generalized set of features.

\begin{table}[h!]
\caption{Generalization - Combine the acoustic features using PCA}
\label{tab_pca_feat}
\setlength{\tabcolsep}{3pt}
\begin{tabular}{|p{100pt}|p{80pt}|p{260pt}|}
\hline
Feature Name & Functional & Principle Component (PC)\\
\hline
MFCC 0 - 14 & mean & 1st PC from the means of 15 MFCCs\\
             &      & 2nd PC from the means of 15 MFCCs \\

             & kurt & 1st PC from the kurt of 15 MFCCs \\
             &          & 2nd PC from the kurt of 15 MFCCs  \\

             & skew & 1st PC from the skew of 15 MFCCs \\
             &          & 2nd PC from the skew of 15 MFCCs  \\
\hline

$\Delta$ MFCC 0 - 14 & mean & 1st PC from the means of 15 $\Delta$ MFCCs\\
                      &      & 2nd PC from the means of 15 $\Delta$ MFCCs \\
                      & kurt & 1st PC from the kurt of 15 $\Delta$ MFCCs \\
                      &          & 2nd PC from the kurt of 15 $\Delta$ MFCCs  \\
                      & skew & 1st PC from the skew of 15 $\Delta$ MFCCs  \\
                      &          & 2nd PC from the skew of 15 $\Delta$ MFCCs   \\
\hline

$\Delta$ LSP freq 0 - 7 & mean & 1st PC from the means of 8 $\Delta$ LSP freq \\
                                  &      & 2nd PC from the means of 8 $\Delta$ LSP freq \\
                                  & kurt & 1st PC from the kurt of 8 $\Delta$ LSP freq \\
                                  &          & 2nd PC from the kurt of 8 $\Delta$ LSP freq \\
                                  & skew & 1st PC from the skew of 8 $\Delta$ LSP freq \\
                                  &          & 2nd PC from the skew of 8 $\Delta$ LSP freq  \\
\hline

\end{tabular}
\end{table}

\begin{table}
\caption{Results obtained by applying ML algorithms on PCA-based acoustic features that are extracted from all datasets}
\label{tab_pca_rslt}
\setlength{\tabcolsep}{6pt}
\begin{tabular}{|p{80pt}|p{80pt}|p{80pt}|p{80pt}|p{80pt}|p{80pt}|}
\hline
Classifier & PD &  SR & Web & Phone \\

\hline
DT & \textbf{0.78 (+/- 0.16)} & \textbf{0.78 (+/- 0.02)} & 0.72 (+/- 0.21) & 0.80 (+/- 0.16)\\ 
ET & 0.61 (+/- 0.28) & 0.61 (+/- 0.28) & 0.68 (+/- 0.22) & 0.92 (+/- 0.10) \\  
kNN & 0.61 (+/- 0.08) & \textbf{0.78 (+/- 0.02)}  & 0.61 (+/- 0.08)  & 0.80 (+/- 0.28) \\ 
LDA & 0.50 (+/- 0.14) & 0.50 (+/- 0.14) & 0.50 (+/- 0.14) & 0.87 (+/- 0.09) \\ 
R\_SVM & 0.61 (+/- 0.08) & 0.65 (+/- 0.18) & 0.72 (+/- 0.21) & 0.76 (+/- 0.17) \\ 
L\_SVM & 0.50 (+/- 0.14) & 0.35 (+/- 0.33) & \textbf{0.89 (+/- 0.16)} & 0.67 (+/- 0.25) \\ 
LR & 0.72 (+/- 0.21) & 0.22 (+/- 0.16) & \textbf{0.89 (+/- 0.16)} & 0.60 (+/- 0.28)  \\ 
RF & 0.54 (+/- 0.15) & \textbf{0.78 (+/- 0.02} & 0.79 (+/- 0.23) &  \textbf{0.93 (+/- 0.07)} \\ 
\hline
\end{tabular}
\end{table}

\subsection{Is an AI-powered LA tool Valid and Reliable?}
We refer to reliability as the measure to trust the classification results \cite{kukar2002reliable}. We can assess the reliability and validity of an AI-powered LA tool measuring the \textit{Intra-class Correlation Coefficient} (ICC) \cite{molodynski2013reliability} and the \textit{Pearson Correlation Coefficient}~(PCC) \cite{molodynski2013reliability} (i.e., if PCC returns a value close to 1, then the ML tool provides valid results; however, if the value is below, 0.5 indicates less correlation and validation). 
\subsection{Is an AI-powered LA tool Fair?}
AI-powered LA tools are supervised classifiers, and are therefore prone to producing unfair results. In our work, we tried not to consider sensitive attributes such as gender, race as features \cite{barocas2017fairness}. However, we are working on different verbal tasks that might slightly be influenced by gender differences \cite{scheuringer2017sex}. Another issue is that this type of assessment tool compares the user's language against similar users who are assumed to have AD or MCI \cite{9018280}. Another essential attribute that might affect the fairness of AI-powered LA tools is the level of education. It has been shown that some LA tools cannot provide accurate diagnostic when there are subjects with low levels of education among the population of study~\cite{10.3389/fpsyt.2019.00878}. AI-powered LA tools require a set of mechanisms to ensure that end-users trust in their performances and know how the system provides output. It is essential to motivate people to adopt not only the methods but also to share their data.  
Fairness and explainability are essential concerns, especially as AI-powered assessment tools are being deployed more broadly in detecting other types of mental health problems. 
Fairness, in the end, comes down to robustness aspect. When we create our AI-powered assessment tool, we want it to be fair, and this means robust when deployed in different geographic settings and populations. 
\subsection{Is an AI-powered LA tool Explainable?}
An AI-powered LA tool should be accurate and explainable to be adopted by psychiatrists during their assessment procedures. Thus, it is essential to choose an ML algorithm to develop a reliable AI-powered LA tool that can describe its purpose, rationale, and decision-making process that can be understood by both clinicians and patients; it can foster the confidence of mental health professionals in employing it to detect subjects with dementia quickly.  

\subsection{What are other Application of an AI-powered LA tool?}
Automated analysis of spontaneous connected speech can be useful for assessing and monitoring the progress of AD in patients. For example, integrating the AI-powered LA tools to a conversational robot that can record patients' speech provides us, clinicians, an automated approach to follow the progress of the diseases~\cite
{Pou-Prom2020Conversational}. In more detail, we can ask elderly individuals to describe the cookie theft picture to engage them to attend in a conversation. By extracting linguistic and acoustic features from the speech produced by them, we can identify if they are suffering from the linguistic disorder associated with AD or MCI~\cite{Pou-Prom2020Conversational}. 
The AI-powered LA tools are the core part of any smartphone application that aims to support elderly individuals with limited access to clinical services to receive real-time, cost-effective health care services. It decreases the burden on the caregivers.
\section{Conclusion}
In this paper, we suggested an approach to develop an efficient AI-powered language assessment tool. Such a tool can accurately and quickly detect language impairment in the older adults. We showed that 
the assessment tool could be developed using  traditional ML classifiers, which could be trained with various sets of linguistic and acoustic features. 
Our results showed that the classifiers that have been trained using the PD dataset perform better than the SR dataset. We also found that the dataset obtained using phone-based recordings could increase ML classifiers' performance compared to the web-based dataset. Finally, we revealed that the classifiers trained only with the selected features using feature selection methods had higher performance than classifiers trained with the whole set of extracted features.

%This paper suggested a methodology to develop an efficient ML-based language assessment tool to detect language impairment in the elderly. We showed that the 
%assessment tool is compatible with traditional ML classifiers, which can be trained with various sets of linguistic and acoustic features. Even though assessing dementia using ML-based language assessment tools is widely explored in the literature; however, they rarely investigated how different language tasks, features,  and recording media impact the classifiers' performance. Our results showed that the classifiers that have been trained using the PD dataset perform better than the SR dataset. We also found that the dataset obtained using phone-based recordings could increase ML classifiers' performance compared to the web-based dataset. Finally, we revealed that the classifiers trained only with the selected acoustic features using feature selection methods had higher performance than classifiers trained with pure linguistic features. 

In the future, we will be working in the following directions:
1) Developing a cascade classifier that will be trained using both linguistic and acoustic features;
2) Using other types of data, such as eye-tracking; 
2) Using few-shot ML algorithms and transfer learning techniques; 
3) Considering pragmatic features such as fillers, GoAhead utterances, repetitions, incomplete words, and also contextual features using BERT (Bidirectional Encoder Representations from Transformers);
4) Using text data augmentation techniques such as EDA: Easy Data Augmentation techniques to augment data samples.  
5) Classifying data.

\section{Declarations}
\subsection{Consent for publication}
It is not applicable to our manuscript
\subsection{Ethics approval and consent to participate
}
The consent form has been approved by the Research Ethics Board protocol 31127 of the University of Toronto and has been signed by each subject has signed a consent form that has been. 
%\subsection{Consent for publication}
\subsection{Availability of data and material}
The datasets and codes would be accessible upon sending requests to the first author.
\subsection{Funding}
The first author would like to thank The Michael J. Fox Foundation for Parkinson's Research for funding her postdoctoral research projects.
The second and third authors would like to thank AGE-WELL NCE for funding this research. 
%\section*{Competing interests}
  
\section*{Author's contributions}

Dr. Parsa and Dr. Alam worked with the linguistic and acoustic features, respectively, and they contributed equally to the methodology and results in sections.  Dr. Mihailidis reviewed the manuscript and provided valuable feedbacks on the manuscript.  
    
\section*{Acknowledgements}
The first and second authors would like to thank Dr. Frank Rudzicz for providing access to the Talk2Me Database and valuable academic suggestions. The first author would like to thank Mrs. Marina Tawfik for extracting datasets from the Talk2Me DB and reviewing the paper. 

%\section{Headings: first level}
%\label{sec:headings}

%Bibliography
\bibliographystyle{unsrt}  
\bibliography{references}

\begin{thebibliography}{10}

\bibitem{ripich2004neurodegenerative}
Danielle~N Ripich and Jennifer Horner.
\newblock The neurodegenerative dementias: Diagnoses and interventions.
\newblock {\em The ASHA Leader}, 9(8):4--15, 2004.

\bibitem{nichols2019global}
Emma Nichols, Cassandra~EI Szoeke, Stein~Emil Vollset, Nooshin Abbasi, Foad
  Abd-Allah, Jemal Abdela, Miloud Taki~Eddine Aichour, Rufus~O Akinyemi, Fares
  Alahdab, Solomon~W Asgedom, et~al.
\newblock Global, regional, and national burden of alztteimer's disease and
  other dementias, 1990--2016: a systematic analysis for the global burden of
  disease study 2016.
\newblock {\em The Lancet Neurology}, 18(1):88--106, 2019.

\bibitem{santacruz2001early}
Karen SantaCruz and Daniel~L Swagerty~Jr.
\newblock Early diagnosis of dementia.
\newblock {\em American Family Physician}, 63(4):703, 2001.

\bibitem{green1997early}
RC~Green, VC~Clarke, NJ~Thompson, JL~Woodard, and R~Letz.
\newblock Early detection of alzheimer disease: methods, markers, and
  misgivings.
\newblock {\em Alzheimer disease and associated disorders}, 11(5):S1, 1997.

\bibitem{duan2018psychosocial}
Yuting Duan, Liming Lu, Juexuan Chen, Chunxiao Wu, Jielin Liang, Yan Zheng,
  Jinjian Wu, Peijing Rong, and Chunzhi Tang.
\newblock Psychosocial interventions for alzheimer’s disease cognitive
  symptoms: a bayesian network meta-analysis.
\newblock {\em BMC geriatrics}, 18(1):175, 2018.

\bibitem{fischer2019music}
Corinne~Eleanor Fischer.
\newblock Music intervention approaches for alzheimer’s disease: a review of
  the literature.
\newblock {\em Frontiers in Neuroscience}, 13:132, 2019.

\bibitem{logsdon2007evidence}
Rebecca~G Logsdon, Susan~M McCurry, and Linda Teri.
\newblock Evidence-based interventions to improve quality of life for
  individuals with dementia.
\newblock {\em Alzheimer's care today}, 8(4):309, 2007.

\bibitem{10.1001/jamainternmed.2015.2152}
Kelvin K.~F. Tsoi, Joyce Y.~C. Chan, Hoyee~W. Hirai, Samuel Y.~S. Wong, and
  Timothy C.~Y. Kwok.
\newblock {Cognitive Tests to Detect Dementia: A Systematic Review and
  Meta-analysis}.
\newblock {\em JAMA Internal Medicine}, 175(9):1450--1458, 09 2015.

\bibitem{creavin2016mini}
Sam~T Creavin, Susanna Wisniewski, Anna~H Noel-Storr, Clare~M Trevelyan, Thomas
  Hampton, Dane Rayment, Victoria~M Thom, Kirsty~JE Nash, Hosam Elhamoui,
  Rowena Milligan, et~al.
\newblock Mini-mental state examination (mmse) for the detection of dementia in
  clinically unevaluated people aged 65 and over in community and primary care
  populations.
\newblock {\em Cochrane Database of Systematic Reviews}, (1), 2016.

\bibitem{10.3389/fpsyt.2019.00878}
José Wagner~Leonel Tavares-Júnior, Ana Célia~Caetano de~Souza,
  Gilberto~Sousa Alves, Janine de~Carvalho Bonfadini, José~Ibiapina
  Siqueira-Neto, and Pedro Braga-Neto.
\newblock Cognitive assessment tools for screening older adults with low levels
  of education: A critical review.
\newblock {\em Frontiers in Psychiatry}, 10:878, 2019.

\bibitem{daly2005initial}
Mel~P Daly.
\newblock Initial evaluation of the patient with suspected dementia.
\newblock {\em American Family Physician}, 71(9):1745--1750, 2005.

\bibitem{kalish2016mini}
Virginia~B Kalish and Brian Lerner.
\newblock Mini-mental state examination for the detection of dementia in older
  patients.
\newblock {\em American family physician}, 94(11):880--881, 2016.

\bibitem{nasreddine2005montreal}
Ziad~S Nasreddine, Natalie~A Phillips, Val{\'e}rie B{\'e}dirian, Simon
  Charbonneau, Victor Whitehead, Isabelle Collin, Jeffrey~L Cummings, and
  Howard Chertkow.
\newblock The montreal cognitive assessment, moca: a brief screening tool for
  mild cognitive impairment.
\newblock {\em Journal of the American Geriatrics Society}, 53(4):695--699,
  2005.

\bibitem{chiu2019nmd}
Pai-Yi Chiu, Haipeng Tang, Cheng-Yu Wei, Chaoyang Zhang, Guang-Uei Hung, and
  Weihua Zhou.
\newblock Nmd-12: A new machine-learning derived screening instrument to detect
  mild cognitive impairment and dementia.
\newblock {\em PloS one}, 14(3), 2019.

\bibitem{o2005use}
Shaun~T O'Keeffe, Eamon~C Mulkerrin, Kayser Nayeem, Matthew Varughese, and
  Isweri Pillay.
\newblock Use of serial mini-mental state examinations to diagnose and monitor
  delirium in elderly hospital patients.
\newblock {\em Journal of the American Geriatrics Society}, 53(5):867--870,
  2005.

\bibitem{vertesi2001standardized}
Andrea Vertesi, Judith~A Lever, D~William Molloy, Brett Sanderson, Irene
  Tuttle, Laura Pokoradi, and Elaine Principi.
\newblock Standardized mini-mental state examination. use and interpretation.
\newblock {\em Canadian Family Physician}, 47(10):2018--2023, 2001.

\bibitem{sheehan2012assessment}
Bart Sheehan.
\newblock Assessment scales in dementia.
\newblock {\em Therapeutic advances in neurological disorders}, 5(6):349--358,
  2012.

\bibitem{chambers2017dementia}
Larry~W Chambers, Saskia Sivananthan, and Carol Brayne.
\newblock Is dementia screening of apparently healthy individuals justified?
\newblock {\em Advances in preventive medicine}, 2017, 2017.

\bibitem{chaves2011cognitive}
M{\'a}rcia~LF Chaves, Claudia~C Godinho, Claudia~S Porto, Leticia Mansur,
  Maria~Teresa Carthery-Goulart, M{\^o}nica~S Yassuda, and Rog{\'e}rio Beato.
\newblock Cognitive, functional and behavioral assessment: Alzheimer's disease.
\newblock {\em Dementia \& neuropsychologia}, 2011.

\bibitem{el2018novel}
Fatma~EA El-Gamal, Mohammed~M Elmogy, Mohammed Ghazal, Ahmed Atwan, Manuel~F
  Casanova, Gregory~N Barnes, Robert Keynton, Ayman~S El-Baz, and Ashraf
  Khalil.
\newblock A novel early diagnosis system for mild cognitive impairment based on
  local region analysis: A pilot study.
\newblock {\em Frontiers in human neuroscience}, 11:643, 2018.

\bibitem{klimova2015alzheimer}
Blanka Klimova, Petra Maresova, Martin Valis, Jakub Hort, and Kamil Kuca.
\newblock Alzheimer’s disease and language impairments: social intervention
  and medical treatment.
\newblock {\em Clinical interventions in aging}, 10:1401, 2015.

\bibitem{doi:10.1111/hsc.12887}
Luisa Krein, Yun-Hee Jeon, and Amanda Miller~Amberber.
\newblock Development of a new tool for the early identification of
  communication-support needs in people living with dementia: An australian
  face-validation study.
\newblock {\em Health \& Social Care in the Community}, 28(2):544--554, 2020.

\bibitem{green2018investigating}
Samantha Green, Satu Reivonen, Lisa-Marie Rutter, Eva Nouzova, Nikki Duncan,
  Caoimhe Clarke, Alasdair~MJ MacLullich, and Zoe Tieges.
\newblock Investigating speech and language impairments in delirium: a
  preliminary case-control study.
\newblock {\em PloS one}, 13(11), 2018.

\bibitem{szatloczki2015speaking}
Greta Szatloczki, Ildiko Hoffmann, Veronika Vincze, Janos Kalman, and Magdolna
  Pakaski.
\newblock Speaking in alzheimer’s disease, is that an early sign? importance
  of changes in language abilities in alzheimer’s disease.
\newblock {\em Frontiers in aging neuroscience}, 7:195, 2015.

\bibitem{ferreira2011neuroimaging}
Luiz~Kobuti Ferreira and Geraldo~F Busatto.
\newblock Neuroimaging in alzheimer's disease: current role in clinical
  practice and potential future applications.
\newblock {\em Clinics}, 66:19--24, 2011.

\bibitem{mccullough2019language}
Kim~C McCullough, Kathryn~A Bayles, and Erin~D Bouldin.
\newblock Language performance of individuals at risk for mild cognitive
  impairment.
\newblock {\em Journal of Speech, Language, and Hearing Research},
  62(3):706--722, 2019.

\bibitem{vestal2006efficacy}
Lindsey Vestal, Laura Smith-Olinde, Gretchen Hicks, Terri Hutton, and John
  Hart~Jr.
\newblock Efficacy of language assessment in alzheimer’s disease: comparing
  in-person examination and telemedicine.
\newblock {\em Clinical interventions in aging}, 1(4):467, 2006.

\bibitem{godino2005support}
Juan~Ignacio Godino-Llorente, Pedro G{\'o}mez-Vilda, Nicol{\'a}s
  S{\'a}enz-Lech{\'o}n, Manuel Blanco-Velasco, Fernando Cruz-Rold{\'a}n, and
  Miguel~Angel Ferrer-Ballester.
\newblock Support vector machines applied to the detection of voice disorders.
\newblock In {\em International Conference on Nonlinear Analyses and Algorithms
  for Speech Processing}, pages 219--230. Springer, 2005.

\bibitem{guinn2012language}
Curry~I Guinn and Anthony Habash.
\newblock Language analysis of speakers with dementia of the alzheimer’s
  type.
\newblock In {\em 2012 AAAI Fall Symposium Series}, 2012.

\bibitem{orimaye-etal-2014-learning}
Sylvester~Olubolu Orimaye, Jojo Sze-Meng Wong, and Karen~Jennifer Golden.
\newblock Learning predictive linguistic features for {A}lzheimer{'}s disease
  and related dementias using verbal utterances.
\newblock In {\em Proceedings of the Workshop on Computational Linguistics and
  Clinical Psychology: From Linguistic Signal to Clinical Reality}, pages
  78--87, Baltimore, Maryland, USA, June 2014. Association for Computational
  Linguistics.

\bibitem{asgari2017predicting}
Meysam Asgari, Jeffrey Kaye, and Hiroko Dodge.
\newblock Predicting mild cognitive impairment from spontaneous spoken
  utterances.
\newblock {\em Alzheimer's \& Dementia: Translational Research \& Clinical
  Interventions}, 3(2):219--228, 2017.

\bibitem{karlekar2018detecting}
Sweta Karlekar, Tong Niu, and Mohit Bansal.
\newblock Detecting linguistic characteristics of alzheimer's dementia by
  interpreting neural models.
\newblock {\em arXiv preprint arXiv:1804.06440}, 2018.

\bibitem{pope2011finding}
Charlene Pope and Boyd~H Davis.
\newblock Finding a balance: The carolinas conversation collection.
\newblock {\em Corpus Linguistics and Linguistic Theory}, 7(1):143--161, 2011.

\bibitem{becker1994natural}
James~T Becker, Fran{\c{c}}ois Boiler, Oscar~L Lopez, Judith Saxton, and
  Karen~L McGonigle.
\newblock The natural history of alzheimer's disease: description of study
  cohort and accuracy of diagnosis.
\newblock {\em Archives of Neurology}, 51(6):585--594, 1994.

\bibitem{calero2002usefulness}
M~Dolores Calero, M~Luisa Arnedo, Elena Navarro, M{\'o}nica Ruiz-Pedrosa, and
  Cristobal Carnero.
\newblock Usefulness of a 15-item version of the boston naming test in
  neuropsychological assessment of low-educational elders with dementia.
\newblock {\em The Journals of Gerontology Series B: Psychological Sciences and
  Social Sciences}, 57(2):P187--P191, 2002.

\bibitem{slegers2018connected}
Antoine Slegers, Ren{\'e}e-Pier Filiou, Maxime Montembeault, and Simona~Maria
  Brambati.
\newblock Connected speech features from picture description in alzheimer’s
  disease: A systematic review.
\newblock {\em Journal of Alzheimer's Disease}, (Preprint):1--26, 2018.

\bibitem{cummings2019describing}
Louise Cummings.
\newblock Describing the cookie theft picture: Sources of breakdown in
  alzheimer’s dementia.
\newblock {\em Pragmatics and Society}, 10(2):153--176, 2019.

\bibitem{weissenbacher2016automatic}
Davy Weissenbacher, Travis~A Johnson, Laura Wojtulewicz, Amylou Dueck, Dona
  Locke, Richard Caselli, and Graciela Gonzalez.
\newblock Automatic prediction of linguistic decline in writings of subjects
  with degenerative dementia.
\newblock In {\em Proceedings of the 2016 Conference of the North American
  Chapter of the Association for Computational Linguistics: Human Language
  Technologies}, pages 1198--1207, 2016.

\bibitem{araujo2015linguagem}
Aline Menezes Guedes Dias~de Ara{\'u}jo, Daviany~Oliveira Lima, Islan da~Penha
  Nascimento, Anna Alice Figueir{\^e}do~de Almeida, and Marine Raquel Diniz~da
  Rosa.
\newblock Linguagem em idosos com doen{\c{c}}a de alzheimer: uma revis{\~a}o
  sistem{\'a}tica.
\newblock {\em Revista CEFAC}, 17(5):1657--1663, 2015.

\bibitem{orimaye2017predicting}
Sylvester~O Orimaye, Jojo~SM Wong, Karen~J Golden, Chee~P Wong, and Ireneous~N
  Soyiri.
\newblock Predicting probable alzheimer’s disease using linguistic deficits
  and biomarkers.
\newblock {\em BMC bioinformatics}, 18(1):34, 2017.

\bibitem{Habash2011}
Anthony Habash.
\newblock Language analysis of speakers with dementia of the alzheimer's type,
  2011.

\bibitem{Yancheva2015using}
Maria Yancheva, Kathleen~C Fraser, and Frank Rudzicz.
\newblock Using linguistic features longitudinally to predict clinical scores
  for alzheimer’s disease and related dementias.
\newblock In {\em Proceedings of SLPAT 2015: 6th Workshop on Speech and
  Language Processing for Assistive Technologies}, pages 134--139, 2015.

\bibitem{fraser2016linguistic}
Kathleen~C Fraser, Jed~A Meltzer, and Frank Rudzicz.
\newblock Linguistic features identify alzheimer’s disease in narrative
  speech.
\newblock {\em Journal of Alzheimer's Disease}, 49(2):407--422, 2016.

\bibitem{hernandez2018computer}
Laura Hern{\'a}ndez-Dom{\'\i}nguez, Sylvie Ratt{\'e}, Gerardo
  Sierra-Mart{\'\i}nez, and Andr{\'e}s Roche-Bergua.
\newblock Computer-based evaluation of alzheimer’s disease and mild cognitive
  impairment patients during a picture description task.
\newblock {\em Alzheimer's \& Dementia: Diagnosis, Assessment \& Disease
  Monitoring}, 10:260--268, 2018.

\bibitem{Roark2011spoken}
Brian Roark, Margaret Mitchell, John-Paul Hosom, Kristy Hollingshead, and
  Jeffrey Kaye.
\newblock Spoken language derived measures for detecting mild cognitive
  impairment.
\newblock {\em IEEE transactions on audio, speech, and language processing},
  19(7):2081--2090, 2011.

\bibitem{konig2015automatic}
Alexandra K{\"o}nig, Aharon Satt, Alexander Sorin, Ron Hoory, Orith
  Toledo-Ronen, Alexandre Derreumaux, Valeria Manera, Frans Verhey, Pauline
  Aalten, Phillipe~H Robert, et~al.
\newblock Automatic speech analysis for the assessment of patients with
  predementia and alzheimer's disease.
\newblock {\em Alzheimer's \& Dementia: Diagnosis, Assessment \& Disease
  Monitoring}, 1(1):112--124, 2015.

\bibitem{Ahmed2013connected}
Samrah Ahmed, Anne-Marie~F Haigh, Celeste~A de~Jager, and Peter Garrard.
\newblock Connected speech as a marker of disease progression in autopsy-proven
  alzheimer’s disease.
\newblock {\em Brain}, 136(12):3727--3737, 2013.

\bibitem{Meilan2012acoustic}
Juan~JG Meil{\'a}n, Francisco Mart{\'\i}nez-S{\'a}nchez, Juan Carro, Jos{\'e}~A
  S{\'a}nchez, and Enrique P{\'e}rez.
\newblock Acoustic markers associated with impairment in language processing in
  alzheimer's disease.
\newblock {\em The Spanish journal of psychology}, 15(2):487--494, 2012.

\bibitem{pasrapoor2013emotional}
Mahboobeh Pasrapoor and Urban Bilstrup.
\newblock An emotional learning-inspired ensemble classifier (eliec).
\newblock In {\em 2013 Federated Conference on Computer Science and Information
  Systems}, pages 137--141. IEEE, 2013.

\bibitem{Parsa-2020_Alzheimer}
Mahboobeh Parsapoor.
\newblock {Detecting language impairment using ELIEC}.
\newblock In Alzheimer's Association, editor, {\em {DEMENTIA CARE AND
  PSYCHOSOCIAL FACTORS}}, volume~16, page e046767, 2020.

\bibitem{noorian2017importance}
Zeinab Noorian, Chlo{\'e} Pou-Prom, and Frank Rudzicz.
\newblock On the importance of normative data in speech-based assessment.
\newblock {\em arXiv preprint arXiv:1712.00069}, 2017.

\bibitem{Loper02nltk:the}
Edward Loper and Steven Bird.
\newblock Nltk: The natural language toolkit.
\newblock In {\em In Proceedings of the ACL Workshop on Effective Tools and
  Methodologies for Teaching Natural Language Processing and Computational
  Linguistics. Philadelphia: Association for Computational Linguistics}, 2002.

\bibitem{malvern2004lexical}
David Malvern, Brian Richards, Ngoni Chipere, and Pilar Dur{\'a}n.
\newblock {\em Lexical diversity and language development}.
\newblock Springer.

\bibitem{kincaid1975derivation}
J~Peter Kincaid, Robert~P Fishburne~Jr, Richard~L Rogers, and Brad~S Chissom.
\newblock Derivation of new readability formulas (automated readability index,
  fog count and flesch reading ease formula) for navy enlisted personnel.
\newblock 1975.

\bibitem{sakai2011linguistic}
Erin~Y Sakai and Brian~D Carpenter.
\newblock Linguistic features of power dynamics in triadic dementia diagnostic
  conversations.
\newblock {\em Patient education and counseling}, 85(2):295--298, 2011.

\bibitem{komeili2019talk2me}
Majid Komeili, Chlo{\'e} Pou-Prom, Daniyal Liaqat, Kathleen~C Fraser, Maria
  Yancheva, and Frank Rudzicz.
\newblock Talk2me: Automated linguistic data collection for personal
  assessment.
\newblock {\em PloS one}, 14(3), 2019.

\bibitem{peelle2007syntactic}
Jonathan~E Peelle, Ayanna Cooke, Peachie Moore, Luisa Vesely, and Murray
  Grossman.
\newblock Syntactic and thematic components of sentence processing in
  progressive nonfluent aphasia and nonaphasic frontotemporal dementia.
\newblock {\em Journal of Neurolinguistics}, 20(6):482--494, 2007.

\bibitem{Parsa-NN2020}
Mahboobeh Parsapoor.
\newblock {Measuring Tangential Speech in Patients with Dementia}.
\newblock In Alzheimer's Association, editor, {\em {Neuroscience Next 2020
  Abstracts}}, volume~16, page e12278, 2020.

\bibitem{arora2016simple}
Sanjeev Arora, Yingyu Liang, and Tengyu Ma.
\newblock A simple but tough-to-beat baseline for sentence embeddings.
\newblock 2016.

\bibitem{blei2003latent}
David~M Blei, Andrew~Y Ng, and Michael~I Jordan.
\newblock Latent dirichlet allocation.
\newblock {\em Journal of machine Learning research}, 3(Jan):993--1022, 2003.

\bibitem{landauer1998introduction}
Thomas~K Landauer, Peter~W Foltz, and Darrell Laham.
\newblock An introduction to latent semantic analysis.
\newblock {\em Discourse processes}, 25(2-3):259--284, 1998.

\bibitem{https://doi.org/10.48550/arxiv.2009.13602}
Jonathan Smith, Borna Ghotbi, Seungeun Yi, and Mahboobeh Parsapoor.
\newblock Non-pharmaceutical intervention discovery with topic modeling, 2020.

\bibitem{https://doi.org/10.1002/alz.12278}
Alzheimer's Association.
\newblock Neuroscience next 2020 abstracts.
\newblock {\em Alzheimer's \& Dementia}, 16(S12):e12278, 2020.

\bibitem{McLoughlin2008Line}
Ian~Vince McLoughlin.
\newblock Line spectral pairs.
\newblock {\em Signal Processing}, 88(3):448 -- 467, 2008.

\bibitem{mcloughlin1999lsp}
Ian~Vince McLoughlin and Srikanthan Thambipillai.
\newblock Lsp parameter interpretation for speech classification.
\newblock In {\em ICECS'99. Proceedings of ICECS'99. 6th IEEE International
  Conference on Electronics, Circuits and Systems (Cat. No. 99EX357)},
  volume~1, pages 419--422. IEEE, 1999.

\bibitem{de2002yin}
Alain De~Cheveign{\'e} and Hideki Kawahara.
\newblock Yin, a fundamental frequency estimator for speech and music.
\newblock {\em The Journal of the Acoustical Society of America},
  111(4):1917--1930, 2002.

\bibitem{tsanas2011nonlinear}
Athanasios Tsanas, Max~A Little, Patrick~E McSharry, and Lorraine~O Ramig.
\newblock Nonlinear speech analysis algorithms mapped to a standard metric
  achieve clinically useful quantification of average parkinson's disease
  symptom severity.
\newblock {\em Journal of the royal society interface}, 8(59):842--855, 2011.

\bibitem{yanushevskaya2013voice}
Irena Yanushevskaya, Christer Gobl, and Ailbhe N{\'\i}~Chasaide.
\newblock Voice quality in affect cueing: does loudness matter?
\newblock {\em Frontiers in psychology}, 4:335, 2013.

\bibitem{meilan2014speech}
Juan Jos{\'e}~G Meil{\'a}n, Francisco Mart{\'\i}nez-S{\'a}nchez, Juan Carro,
  Dolores~E L{\'o}pez, Lymarie Millian-Morell, and Jos{\'e}~M Arana.
\newblock Speech in alzheimer's disease: Can temporal and acoustic parameters
  discriminate dementia?
\newblock {\em Dementia and Geriatric Cognitive Disorders}, 37(5-6):327--334,
  2014.

\bibitem{lopez2013automatic}
Karmele Lopez-de Ipina, Jesus~Bernardino Alonso, Carlos~M Travieso, Harkaitz
  Egiraun, Miriam Ecay, Aitzol Ezeiza, Nora Barroso, and Pablo Martinez-Lage.
\newblock Automatic analysis of emotional response based on non-linear speech
  modeling oriented to alzheimer disease diagnosis.
\newblock In {\em 2013 IEEE 17th International Conference on Intelligent
  Engineering Systems (INES)}, pages 61--64. IEEE, 2013.

\bibitem{pedregosa2011scikit}
Fabian Pedregosa, Ga{\"e}l Varoquaux, Alexandre Gramfort, Vincent Michel,
  Bertrand Thirion, Olivier Grisel, Mathieu Blondel, Peter Prettenhofer, Ron
  Weiss, Vincent Dubourg, et~al.
\newblock Scikit-learn: Machine learning in python.
\newblock {\em Journal of machine learning research}, 12(Oct):2825--2830, 2011.

\bibitem{domingos2012few}
Pedro Domingos.
\newblock A few useful things to know about machine learning.
\newblock {\em Communications of the ACM}, 55(10):78--87, 2012.

\bibitem{suresh2012sample}
KP~Suresh and S~Chandrashekara.
\newblock Sample size estimation and power analysis for clinical research
  studies.
\newblock {\em Journal of human reproductive sciences}, 5(1):7, 2012.

\bibitem{kukar2002reliable}
Matja{\v{z}} Kukar and Igor Kononenko.
\newblock Reliable classifications with machine learning.
\newblock In {\em European Conference on Machine Learning}, pages 219--231.
  Springer, 2002.

\bibitem{molodynski2013reliability}
Andrew Molodynski, Michael Linden, George Juckel, Ksenija Yeeles, Catriona
  Anderson, Maria Vazquez-Montes, and Tom Burns.
\newblock The reliability, validity, and applicability of an english language
  version of the mini-icf-app.
\newblock {\em Social psychiatry and psychiatric epidemiology},
  48(8):1347--1354, 2013.

\bibitem{barocas2017fairness}
Solon Barocas, Moritz Hardt, and Arvind Narayanan.
\newblock Fairness in machine learning.

\bibitem{scheuringer2017sex}
Andrea Scheuringer, Ramona Wittig, and Belinda Pletzer.
\newblock Sex differences in verbal fluency: the role of strategies and
  instructions.
\newblock {\em Cognitive processing}, 18(4):407--417, 2017.

\bibitem{9018280}
C.~{Burr}, J.~{Morley}, M.~{Taddeo}, and L.~{Floridi}.
\newblock Digital psychiatry: Risks and opportunities for public health and
  wellbeing.
\newblock {\em IEEE Transactions on Technology and Society}, 1(1):21--33, 2020.

\bibitem{Pou-Prom2020Conversational}
Chlo{\'e} Pou-Prom, Stefania Raimondo, and Frank Rudzicz.
\newblock A conversational robot for older adults with alzheimer’s disease.
\newblock {\em ACM Transactions on Human-Robot Interaction (THRI)}, 9(3):1--25,
  2020.

\end{thebibliography}

\end{document}